\documentclass{article} 
\usepackage{iclr2026_conference,times}

\usepackage{amsmath,amsfonts,bm}

\def\eqref#1{equation~\ref{#1}}

\def\1{\bm{1}}

\DeclareMathAlphabet{\mathsfit}{\encodingdefault}{\sfdefault}{m}{sl}
\SetMathAlphabet{\mathsfit}{bold}{\encodingdefault}{\sfdefault}{bx}{n}

\usepackage{graphicx}
\usepackage[table]{xcolor}
\usepackage{wrapfig}
\usepackage{multirow}
\usepackage{pifont}
\usepackage{enumitem}
\usepackage{tabularx}
\usepackage[utf8]{inputenc}
\usepackage[most]{tcolorbox}
\usepackage{listings}
\usepackage{textcomp}
\usepackage{makecell}
\usepackage{booktabs}
\usepackage{amsfonts}
\usepackage{nicefrac}
\usepackage{microtype}

\definecolor{opencolor}{HTML}{2F7F9D}
\definecolor{closecolor}{HTML}{A0A4AA}
\definecolor{cityblue}{RGB}{128, 159, 225}
\definecolor{adcolor}{HTML}{7D3109}

\newcommand{\opencolor}[1]{\textcolor{opencolor}{#1}}
\newcommand{\closecolor}[1]{\textcolor{closecolor}{#1}}
\newcommand{\open}{\opencolor{$\bullet$\,}} 
\newcommand{\close}{\closecolor{$\bullet$\,}} 
\newcommand*\samethanks[1][\value{footnote}]{\footnotemark[#1]}

\let\titleold\title
\renewcommand{\title}[1]{\titleold{#1}\newcommand{\thetitle}{#1}}

\newcommand{\appsection}[3][1em]{%
  \vspace{#1}%
  \noindent
  \hyperref[#2]{\textbf{#3}} \dotfill \pageref{#2} \par
}

\newcommand{\appsubsection}[3][.5em]{%
  \vspace{#1}%
  \noindent
  \hspace{1em} \hyperref[#2]{#3} \dotfill \pageref{#2} \par
}

\newtcolorbox[auto counter, number within=section]{LongJudgeBox}[2][]{
    enhanced,
    breakable=false,
    colback=white,
    colframe=gray!20,
    boxrule=0.8pt,
    arc=2pt,
    fonttitle=\bfseries\sffamily,
    coltitle=black,
    attach boxed title to top left={xshift=3mm, yshift=-3mm},
    boxed title style={colback=gray!10, sharp corners, frame hidden},
    title={#2},
    middle=1.5mm,
    fontupper=\fontfamily{ptm}\selectfont\tiny,
    fontlower=\fontfamily{ptm}\selectfont\tiny,
    segmentation style={dash pattern=on 3pt off 2pt, draw=gray!40},
    before upper={\setlength{\parindent}{0pt}\setlength{\parskip}{0pt}},
    after upper={}, 
    before lower={\setlength{\parindent}{0pt}\setlength{\parskip}{0pt}},
    #1
}

\newtcolorbox[auto counter, number within=section]{JudgeBox}[2][]{
    enhanced,
    breakable=false,
    colback=white,
    colframe=gray!20,
    boxrule=0.8pt,
    arc=2pt,
    fonttitle=\bfseries\sffamily,
    coltitle=black,
    attach boxed title to top left={xshift=3mm, yshift=-3mm},
    boxed title style={colback=gray!10, sharp corners, frame hidden},
    title={#2},
    middle=1.5mm,
    fontupper=\fontfamily{ptm}\selectfont\small,
    fontlower=\fontfamily{ptm}\selectfont\small,
    segmentation style={dash pattern=on 3pt off 2pt, draw=gray!40},
    before upper={\setlength{\parindent}{0pt}\setlength{\parskip}{0pt}},
    after upper={}, 
    before lower={\setlength{\parindent}{0pt}\setlength{\parskip}{0pt}},
    #1
}
\usepackage{hyperref}
\usepackage{url}

\title{GEditBench v2: A Human-Aligned Benchmark for General Image Editing}

\author{%
  Zhangqi Jiang\textsuperscript{1,}\thanks{Work done during an internship at StepFun} \quad Zheng Sun\textsuperscript{2} \quad Xianfang Zeng\textsuperscript{2}\thanks{Project leader} \quad Yufeng Yang\textsuperscript{2} \quad Xuanyang Zhang\textsuperscript{2} \\[.2em] \textbf{Yongliang Wu\textsuperscript{3}} \quad\quad \textbf{Wei Cheng\textsuperscript{2}} \quad\quad \textbf{Gang Yu\textsuperscript{2}} \quad\quad \textbf{Xu Yang\textsuperscript{3}\thanks{Corresponding author}} \quad\quad \textbf{Bihan Wen\textsuperscript{1}\samethanks}\\[.4em]
  {\textsuperscript{1}Nanyang Technological University
  \quad
  \textsuperscript{2}StepFun
  \quad
  \textsuperscript{3}Southeast University}
  \\[.4em]
  \textcolor{cityblue}{
  \raisebox{-0.2\height}{\includegraphics[height=.4cm]{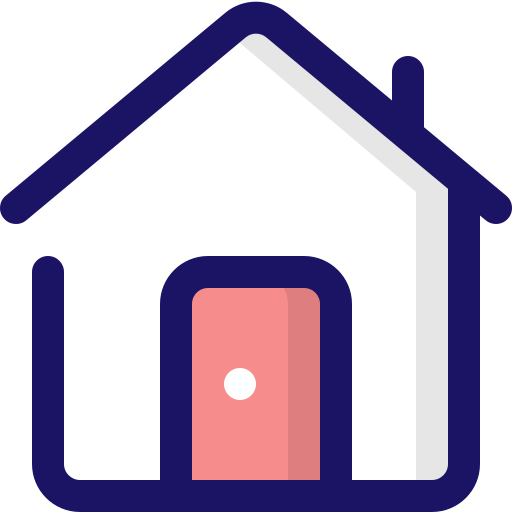}}~{\href{https://zhangqijiang07.github.io/gedit2_web/}{\texttt{\small Project Page}}}
  \quad
  \raisebox{-0.2\height}{\includegraphics[height=.4cm]{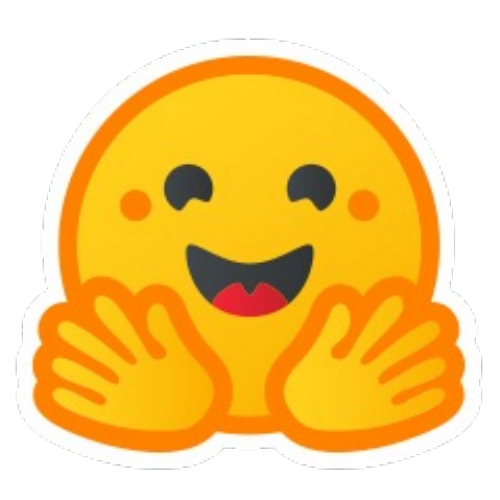}}~{\href{https://huggingface.co/datasets/GEditBench-v2/GEditBench-v2}{\texttt{\small GEditBench v2}}}
  \quad
  \raisebox{-0.2\height}{\includegraphics[height=.4cm]{files/huggingface_logo.pdf}}~{\href{https://huggingface.co/datasets/GEditBench-v2/VCReward-Bench}{\texttt{\small VCReward}}
  \quad
  \raisebox{-0.2\height}{\includegraphics[height=.4cm]{files/huggingface_logo.pdf}}}~{\href{https://huggingface.co/GEditBench-v2/PVC-Judge}{\texttt{\small PVC-Judge}}
  \quad
  \raisebox{-0.2\height}{\includegraphics[height=.4cm]{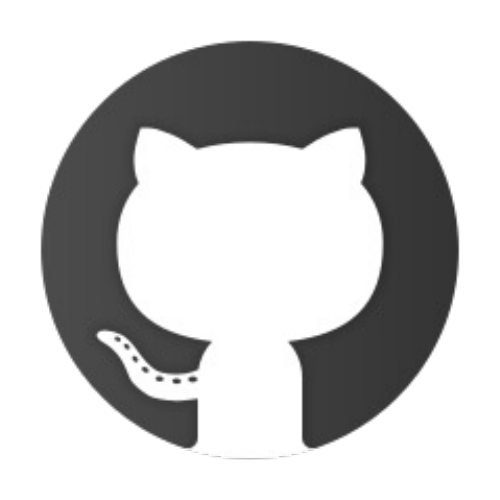}}~{\href{https://github.com/ZhangqiJiang07/GEditBench_v2}{\texttt{\small Code}}}
  }
  }
}

\iclrfinalcopy 
\begin{document}

\maketitle

\begin{abstract}
Recent advances in image editing have enabled models to handle complex instructions with impressive realism.
However, existing evaluation frameworks lag behind: current benchmarks suffer from narrow task coverage, while standard metrics fail to adequately capture \textit{visual consistency}, i.e., the preservation of identity, structure and semantic coherence between edited and original images.
To address these limitations, we introduce \textbf{GEditBench v2}, a comprehensive benchmark with 1,200 real-world user queries spanning 23 tasks, including a dedicated \textit{open-set} category for unconstrained, out-of-distribution editing instructions beyond predefined tasks.
Furthermore, we propose \textbf{PVC-Judge}, an open-source pairwise assessment model for visual consistency, trained via two novel region-decoupled preference data synthesis pipelines.
Besides, we construct \textbf{VCReward-Bench} using expert-annotated preference pairs to assess the alignment of PVC-Judge with human judgments on visual consistency evaluation.
Experiments show that our PVC-Judge achieves state-of-the-art evaluation performance among open-source models and even surpasses GPT-5.1 on average.
Finally, by benchmarking 16 frontier editing models, we show that GEditBench v2 enables more human-aligned evaluation, revealing critical limitations of current models, and providing a reliable foundation for advancing precise image editing.
\end{abstract}
\section{Introduction}\label{sec:intro}
Instruction-based image editing models~\citep{labs2025kontext,liu2025step1x_edit,wu2025qwenimagetechnicalreport,wu2025omnigen2,GLM_Image,team2025longcat_image,seedream2025seedream} have rapidly evolved to execute complex visual modifications directly from natural language instructions.
Recently, Nano Banana Pro~\citep{team2023gemini} emerged as a landmark model, demonstrating robust generalization across diverse instructions while maintaining exercise fine control for high-fidelity results.
The success of Nano Banana Pro has shifted the community's attention from coarse-grained \textbf{instruction following} toward a more nuanced understanding of instruction boundaries~\citep{yin2025reasonedit} -- the delicate line between instruction following (identifying what must be changed) and what we define as \textbf{visual consistency} (the imperative to preserve non-target elements).
For instance, when tasked with ``replacing a subject's cotton shirt with a silk one,'' an excellent editing model must precisely render the new texture and sheen while strictly preserving the subject's identity, the background illumination, and the spatial geometry of the surrounding environment.
Consequently, such precise control over the editing process has become a key indicator of high-quality image editing models.
However, existing evaluation protocols remain inadequate for assessing their visual consistency capability.

To bridge the aforementioned evaluation gap, recent studies have adopted the VLM-as-a-Judge paradigm to assess visual consistency~\citep{ye2025imgedit,ye2025unicedit,luo2025editscore}.
In general, these approaches design prompt templates based on predefined consistency criteria and then query advanced Vision-Language Models (VLMs), such as GPT-4.1~\citep{gpt4o20250325}, to assign an absolute rating score for each edited image.
Although straightforward and easy to implement, this evaluation protocol suffers from three key limitations.
First, it typically relies on closed-source APIs, making results difficult to reproduce and potentially unstable as the underlying models evolve.
Second, replacing these systems with open-source alternatives introduces an accuracy-cost trade-off: smaller models (e.g., 4B/8B) often lack sufficient priors for reliable judgment, whereas larger models incur substantial deployment cost for inference.
Third, the pointwise scoring scheme is poorly aligned with human judgment, which favors pairwise comparison over absolute rating as evidenced in Fig.~\ref{fig:paradigm_ablation}.
Furthermore, existing benchmarks~\citep{labs2025kontext,yu2025anyedit,ye2025imgedit,liu2025step1x_edit,pan2025ice,ye2025unicedit} typically restrict task coverage to a closed set of predefined editing categories, limiting their ability to evaluate the generalization of editing models in open real-world scenarios.

\begin{figure}[t]
    \centering
    \vspace{-1em}
    \includegraphics[width=1\linewidth]{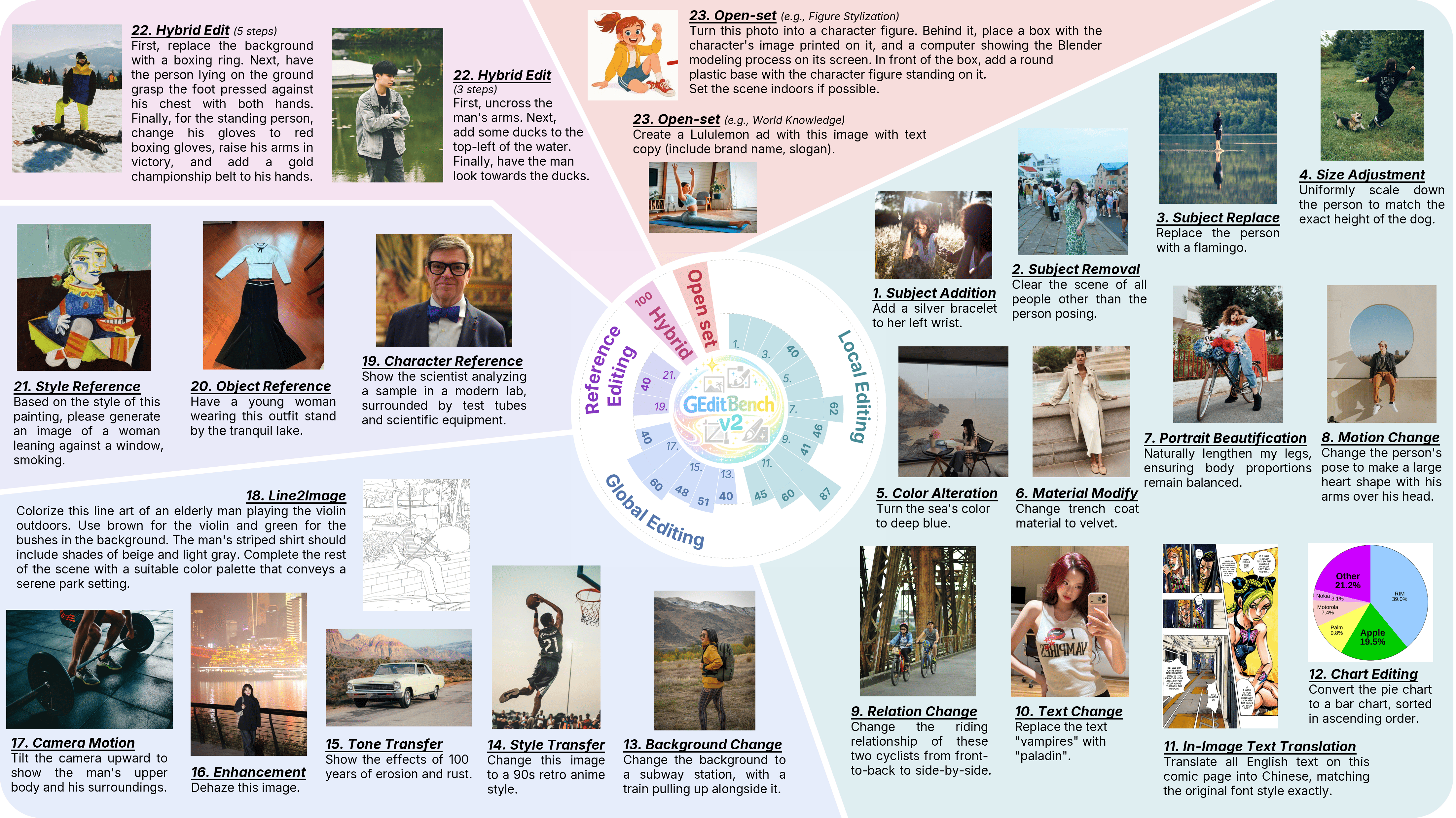}
    \caption{
    \textbf{GEditBench v2} spans 23 diverse image editing tasks, ranging from predefined edits to complex open-set real-world instructions, offering a comprehensive testbed for evaluating instruction-based image editing models.
    The central rose diagram visualizes the corresponding count distribution.
    }
    \label{fig:bench_showcase}
\end{figure}

In this work, we introduce a comprehensive evaluation protocol to address these issues in both model assessment and existing benchmarks.
As shown in Fig.~\ref{fig:bench_showcase}, we first propose \textbf{GEditBench v2}, with an \textit{open-set} category that extends evaluation from standard edit tasks to out-of-distribution instructions, meeting the demands of real-world image editing.
To reliably assess diverse edits and overcome the limitations of the pointwise scheme, we develop \textbf{PVC-Judge}, a human-aligned, \textbf{P}airwise assessment model dedicated to \textbf{V}isual \textbf{C}onsistency.
To train the PVC-Judge, we design novel \textbf{object-} and \textbf{human-centric data curation pipelines} that robustly synthesize high-quality preference pairs at scale by decoupling edited from non-edited regions and ensembling traditional metrics.
Furthermore, to validate the effectiveness of PVC-Judge, we introduce \textbf{VCReward-Bench}, comprising 3,506 expert-annotated preference pairs across 21 predefined tasks, serving as a gold standard for quantifying models' human alignment in assessing visual consistency.
Experimental results on VCReward-Bench show that PVC-Judge achieves the state-of-the-art performance for open-source assessment models, even outperforming GPT-5.1 with an average accuracy of 81.82 versus 76.89.
In summary, our key contributions are as follows:
\begin{itemize}
\item We introduce GEditBench v2, a comprehensive benchmark comprising 22 predefined edit tasks and a dedicated open-set category to evaluate editing models in real-world scenarios.

\item We develop and release PVC-Judge, a pairwise assessment model for visual consistency, trained via two novel region-decoupled preference data synthesis pipelines, achieving human-aligned evaluation.

\item We propose VCReward-Bench, a meta-benchmark to evaluate assessment models for instruction-guided image editing in visual consistency, supported by 3,506 expert-annotated preference pairs.
\end{itemize}
\section{Related Work}
\noindent\textbf{Image Editing Models.}
The field of instruction-based image editing has rapidly evolved from modular, text-guided pipelines~\citep{li2024brushedit,wang2023instructedit} to unified, free-form generative architectures~\citep{zhang2023magicbrush,zhao2024ultraedit,yu2025anyedit}.
Early models like InstructPix2Pix~\citep{brooks2023instructpix2pix} demonstrated the feasibility of diffusion-based editing with synthetic supervision, yet struggled with complex reasoning.
Recent progress addresses this limitation by tightly coupling VLMs with diffusion backbones~\citep{deng2025bagel,wu2025omnigen2,liu2025step1x_edit,team2025longcat_image,wu2025qwenimagetechnicalreport}.
Generally, this integration follows two main paradigms: models such as BAGEL~\citep{deng2025bagel} and OmniGen2~\citep{wu2025omnigen2} jointly optimize multi-modal understanding and generation within a unified framework, whereas decoupled designs like Step1X-Edit~\citep{liu2025step1x_edit} and Qwen-Image-Edit~\citep{wu2025qwenimagetechnicalreport} leverage VLMs as powerful multi-modal encoders to provide structured editing conditions for diffusion transformers.
Concurrently, proprietary systems (e.g., GPT-Image-1.5~\citep{gpt4o20250325}, Nano Banana Pro~\citep{team2023gemini}, and Seedream4.5~\citep{seedream2025seedream}) further advance zero-shot, open-domain editing capabilities through large-scale multi-modal training and
integrated chain-of-thought.
Despite their impressive capabilities, current models still suffer from a fundamental limitation in understanding instruction boundaries, leading to degraded visual consistency.
This gap highlights the need for a rigorous and reliable evaluation of visual consistency in image editing.

\noindent\textbf{Benchmarking for Instruction-based Image Editing.}
Early benchmarking efforts, such as KontextBench~\citep{labs2025kontext}, primarily relied on human evaluation, which is costly and difficult to scale.
Later works like AnyEdit-Bench~\citep{yu2025anyedit} and ICE-Bench~\citep{pan2025ice} introduce automated metrics (e.g., $L_1$-norm, CLIP~\citep{radford2021learning}/DINO~\citep{oquab2023dinov2} scores) for each evaluation dimension, but combining these disparate metrics often leads to fragmented and inconsistent assessments.
Motivated by the VLM-as-a-Judge paradigm~\citep{ku2024viescore}, ImgEdit~\citep{ye2025imgedit}, GEdit~\citep{liu2025step1x_edit} and UnicBench~\citep{ye2025unicedit} benchmarks leverage powerful VLMs (e.g., GPT-4o) to unify evaluation.
However, these approaches remain constrained by their reliance on opaque, closed-source APIs and the use of absolute rating schemes that inherently struggle to capture the relative nature of human preference.
To this end, we develop an 8B assessment model fine-tuned for pairwise comparison, achieving strong human alignment.
In addition, we extend evaluation beyond closed-set task definitions by incorporating open-set instructions derived from trending real-world edits that resist explicit task categorization, enabling a more realistic evaluation of image editing.
A comparative analysis with prior image editing benchmarks is presented in Table~\ref{tab:bench_info}.

\begin{table}[tb]
  \caption{\textbf{Comparison of general image editing benchmarks.}
  GEditBench v2 pioneers a necessary transition to complex open-set scenarios, spanning 23 diverse tasks curated from real-world queries to establish a truly comprehensive evaluation standard.
  }
  \label{tab:bench_info}
  \centering
  \resizebox{\linewidth}{!}{
  \begin{tabular}{lcccccc}
    \toprule
    \multirow{2}{*}{\textbf{Benchmark}} & \multirow{2}{*}{\textbf{Size}} & \multirow{2}{*}{\textbf{Tasks}} & \textbf{Complex Edit} & \textbf{Open-set} & \multicolumn{2}{c}{\textbf{Evaluation}} \\
    & & & \textbf{Support} & \textbf{Instruction} & Unify Metrics & Human-Aligned \\
    \midrule
    
    KontextBench~\cite{labs2025kontext} & 1,026 & 5 & \textcolor{red}{\ding{55}} & \textcolor{red}{\ding{55}} & \textcolor{red}{\ding{55}} & $-$\\
    
    AnyEdit-Bench~\cite{yu2025anyedit} & 1,250 & 25 & \textcolor{red}{\ding{55}} & \textcolor{red}{\ding{55}} & \textcolor{red}{\ding{55}} & Low\\
    
    ImgEdit-Bench~\cite{ye2025imgedit} & 811 & 14 & \textcolor{red}{\ding{55}} & \textcolor{red}{\ding{55}} & \textcolor{green}{\ding{51}} & Mid\\
    
    GEdit-Bench~\cite{liu2025step1x_edit} & 606 & 11 & \textcolor{red}{\ding{55}} & \textcolor{red}{\ding{55}} & \textcolor{green}{\ding{51}} & Mid\\
    
    ICE-Bench~\cite{pan2025ice} & 6,538 & 31 & \textcolor{red}{\ding{55}} & \textcolor{red}{\ding{55}} & \textcolor{red}{\ding{55}} & Low\\
    
    UnicBench~\cite{ye2025unicedit} & 1,100 & 22 & \textcolor{green}{\ding{51}} & \textcolor{red}{\ding{55}} & \textcolor{green}{\ding{51}} & Mid \\
    
    \rowcolor{gray!15}\textbf{GEditBench v2} & 1,200 & 23 & \textcolor{green}{\ding{51}}  & \textcolor{green}{\ding{51}} & \textcolor{green}{\ding{51}} & \textbf{High} \\
  \bottomrule
  \end{tabular}
  }
\end{table}


\section{GEditBench v2}
In this section, we introduce \textbf{GEditBench v2}, a new public benchmark designed to systematically evaluate how the existing editing models can be suffice for user demands in real-world scenarios.

\subsection{Benchmark Construction}\label{subsec:taxonomy}
To ensure comprehensive task coverage, based on~\citep{labs2025kontext,yu2025anyedit,ye2025imgedit,liu2025step1x_edit}, we first structure our benchmark into four main categories, encompassing 19 distinct tasks:
\textbf{1) Local Editing}, which includes edits within a restricted region like \textit{Subject Addition}, \textit{Subject Removal}, \textit{Subject Replace}, \textit{Size Adjustment}, \textit{Color Alteration}, \textit{Material Modification}, \textit{Portrait Beautification}, \textit{Motion Change}, \textit{Relation Change}, and \textit{Text Editing};
\textbf{2) Global Editing}, covering holistic visual transformations such as \textit{Background Change}, \textit{Style Transfer}, \textit{ Tone Transfer}, \textit{Camera Motion}, and \textit{Line2Image};
\textbf{3) Reference Editing}, which tests identity-driven generation such as \textit{Character Reference}, \textit{Object Reference}, and \textit{Style Reference};
\textbf{4) Hybrid Editing}, combining 3$\sim$5 basic edits into a single complex instruction, termed \textit{Hybrid}.

Next, to better reflect real-world user needs within the established taxonomy, we introduce three novel tasks.
Specifically, under the Local Editing category, we introduce \textit{In-Image Text Translation}, which aims to reduce the costs of producing multilingual posters and advertisements, and \textit{Chart Editing}, designed to support chart refinement and chart-type transformation.
Furthermore, within the Global Editing category, we elevate \textit{Enhancement} to an independent task from tone transfer due to its critical practical utility, spanning nine low-level restoration tasks (i.e., blur, compression, moiré, low-light, noise, flare, reflection, haze, and rain), old photo restoration, and overexposed photo rescue.

Finally, to move beyond closed-set paradigms toward real-world, open-ended scenarios, we introduce the fifth category: \textbf{Open-Set Editing}.
This category comprises 100 trending real-world instructions that cannot be explicitly assigned to predefined task taxonomies, enabling a more realistic evaluation of instruction generalization in the wild.

For the above tasks, following~\citep{liu2025step1x_edit}, we collect real-world user editing instances from the Internet, e.g., Reddit and X, and manually filter those editing instructions with a similar intent by trained experts.
To safeguard user privacy, we replace the original user-uploaded images with publicly available images collected from the Internet, supplemented by a small portion generated using Nano Banana Pro~\citep{team2023gemini} or sourced from existing benchmarks~\citep{wu2025omnigen2,liu2025step1x_edit}.
This strategy preserves realistic editing contexts while ensuring privacy protection and reproducibility.
Finally, GEditBench v2 comprises 1,200 testing examples spanning 22 predefined tasks and 1 dedicated open-set editing task.

Notably, we exclude multi-image input editing tasks from our benchmark, as current open-source VLMs exhibit a substantial performance gap from proprietary models in multi-image understanding, making reliable evaluation difficult.
Specifically, according to Table~1 in a recent study~\citep{zhang2025vlm2_bench}, Qwen2.5-VL-7B~\citep{bai2025qwen25vltechnicalreport} underperforms GPT-4o-2024-11-20~\citep{gpt4o20250325} by 8.41\% on average under a four-image setting, with the gap expanding to 30.05\% as the number of input images increases.
Such degradation indicates that existing open-source models are not yet capable of supporting robust multi-image evaluation.
Therefore, this work focuses on single-image editing tasks to ensure accurate assessment.

\begin{figure}[tb]
    \centering
    \includegraphics[width=1\linewidth]{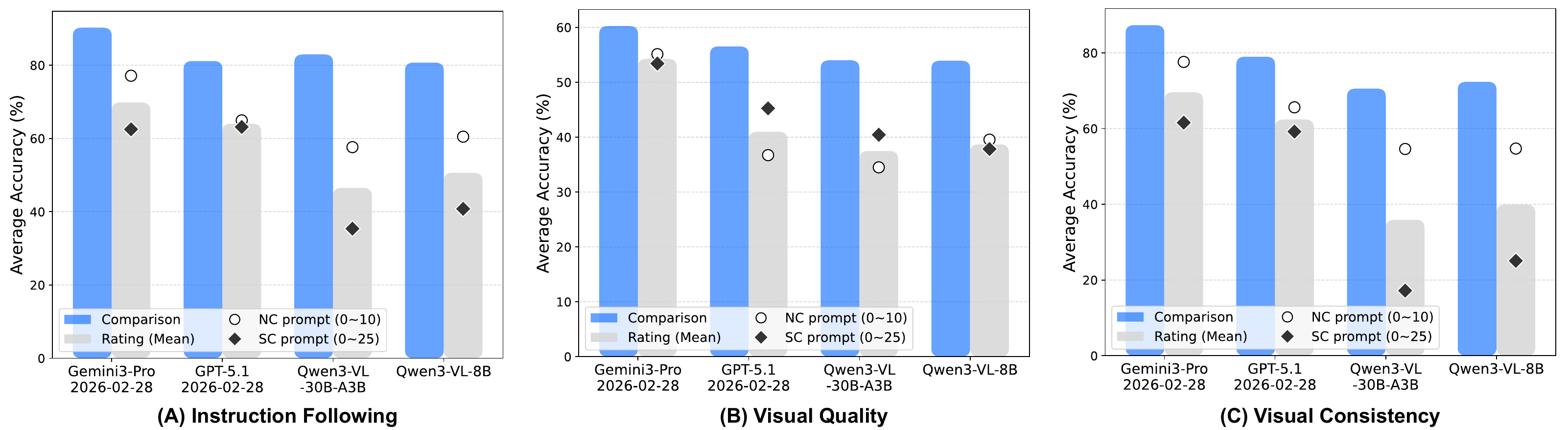}
    \caption{\textbf{Human preference agreement of pointwise and pairwise evaluation paradigms} across four VLMs in (a) instruction following, (b) visual quality, and (c) visual consistency dimensions. Pairwise evaluation consistently achieves higher agreement with human judgments, suggesting its superior human alignment over prior absolute scoring. NC and SC prompts are adopted from~\citep{ye2025unicedit} and~\citep{luo2025editscore}, respectively.}
    \label{fig:paradigm_ablation}
\end{figure}

\subsection{Evaluation Metrics}\label{subsec:eval_metrics}
Following prior works~\citep{liu2025step1x_edit,ye2025imgedit,ye2025unicedit}, we leverage the VLM-as-a-Judge paradigm to evaluate the instruction-based editing models from three dimensions:
\begin{itemize}
\item \textit{Instruction Following} (IF): Measures both prompt comprehension and conceptual understanding of the corresponding prompts.

\item \textit{Visual Quality} (VQ): Evaluates the perceptual quality of the generated image, focusing on overall realism, natural appearance, and the absence of noticeable artifacts.

\item \textit{Visual Consistency} (VC): Assesses preservation of non-target regions, penalizing unintended changes outside the specified edit area.
\end{itemize}

In VLM-as-a-Judge, there are typically two schemes for evaluating generative models: \textit{pointwise rating} and \textit{pairwise comparison}~\citep{chen2024mllm_as_a_judge}.
The pointwise scheme prompts VLMs to assign an absolute score to each image, while the pairwise scheme requires VLMs to express a relative preference between two candidate images.
Despite the wide usage of the pointwise scheme in existing image editing benchmarks~\citep{ye2025imgedit,ye2025unicedit,liu2025step1x_edit,luo2025editscore}, we find that the pairwise scheme is preferable for two key reasons.

\noindent\textbf{1) Stronger Human Alignment:}
To empirically validate this, we evaluated four VLMs across IF, VQ, and VC dimensions on two EditReward-Bench from~\citep{luo2025editscore,wu2025editreward}, randomly swapping image positions with a 50\% probability to mitigate position bias.
The used pointwise prompt templates for visual consistency evaluation, i.e., NC and SC, were proposed in~\citep{ye2025unicedit} and~\citep{luo2025editscore}, respectively.
As shown in Fig.~\ref{fig:paradigm_ablation}, pairwise comparison consistently achieves substantially higher agreement with human judgments across all dimensions, indicating that pairwise preference modeling better reflects human judgment than pointwise rating.

\noindent\textbf{2) Ceiling Effect Mitigation:}
From a training perspective, pointwise evaluators learn a rigid mapping to absolute scores, severely bottlenecking their cognitive upper bound to the training distribution.
When evaluating out-of-distribution edits, they tend to produce similar scores, resulting in a performance ceiling.
Conversely, pairwise training optimizes for relative preference, ensuring robust generalization to new models without losing discriminative power.

Moreover, although pairwise comparison incurs an initial $\mathcal{O}(n^2)$ cost for $n$ candidate models, this is a one-time overhead.
Once the reference pool is established, evaluating a new model requires merely $\mathcal{O}(n)$ comparisons, making the scheme practical and scalable in real-world benchmarking.

Therefore, we adopt pairwise comparison for all evaluation dimensions.
Specifically, for IF, evaluation requires extensive world knowledge to handle diverse and flexible user editing instructions.
Open-source models are generally insufficient for this task, so we rely on GPT-4o~\citep{gpt4o20250325} to perform pairwise assessments.
For VQ, evaluation is instruction-free and can leverage existing text-to-image generation assessment models~\citep{wang2025unified,wu2025visualquality,wang2025pref}.
For simplicity, we also use GPT-4o in a pairwise manner for VQ evaluation.
For VC, to address the limitations discussed in Sec.~\ref{sec:intro} -- reproducibility issues, model size trade-off, and absolute scoring ceiling -- we develop \textbf{PVC-Judge}, an open-source assessment model explicitly fine-tuned for pairwise evaluation of visual consistency in image editing.
Detailed description of PVC-Judge is provided in Sec.~\ref{sec:PVC_Judge}, while the pairwise prompts used for evaluation are presented in the Appendix~\ref{subsec:prompts}.
\section{Pairwise Visual Consistency Judge}\label{sec:PVC_Judge}

\begin{wrapfigure}{r}{0.4\textwidth}
\vspace{-1em}
\centering
\includegraphics[width=\linewidth]{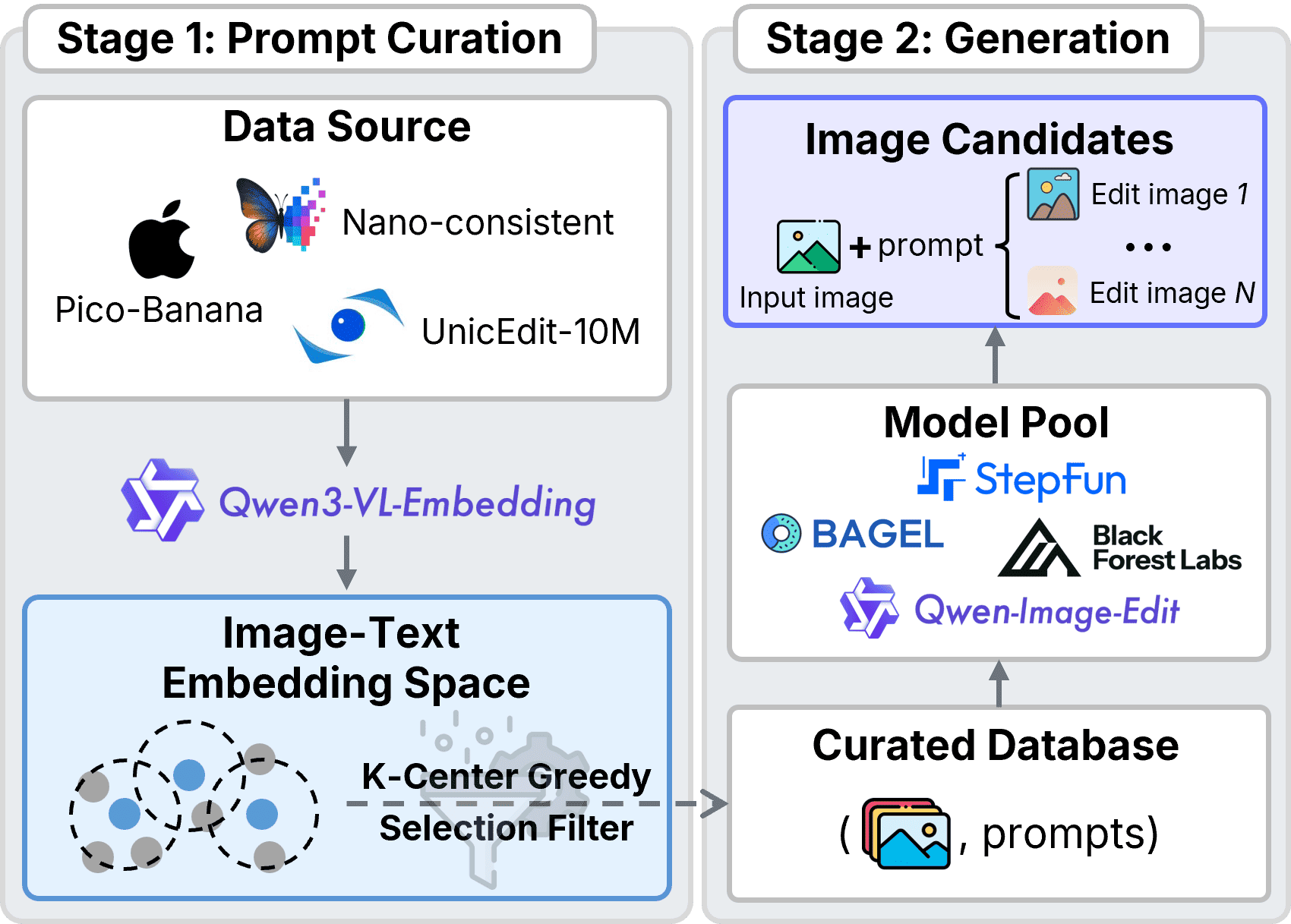}
\caption{Two-stage candidates curation pipeline with prompt filtering.}
\label{fig:candidate_gen}
\vspace{-3em}
\end{wrapfigure}
In this section, we present \textbf{PVC-Judge}, our evaluator for visual consistency in image editing, along with its development pipeline. We first describe candidate image generation (Sec.~\ref{subsec:image_gen}) and two preference data construction pipelines (Sec.~\ref{subsec:preference_data}), followed by the training configuration (Sec.~\ref{subsec:training}). Finally, we introduce \textbf{VCReward-Bench} (Sec.~\ref{subsec:vcreward_bench}), a meta-benchmark for rigorously evaluating the effectiveness of our PVC-Judge.

\subsection{Candidate Image Generation Pipeline}\label{subsec:image_gen}

Before constructing preference pairs, we first build a diverse candidate pool of edited images that enables meaningful comparison.
As shown in Fig.~\ref{fig:candidate_gen}, our pipeline consists of two stages: prompt curation and image generation.
In the first stage, we collect (Input Image, Instruction) pairs, denoted as $(I_{in}, Inst)$, aligned with GEditBench v2's taxonomy from three open-source datasets: Pico-Banana-400K~\citep{qian2025pico}, Nano-Consistency-150K~\citep{ye2025echo}, and UnicEdit-10M~\citep{ye2025unicedit}.
Due to task coverage mismatch, valid pairs are obtained for all tasks except in-image text translation, relation change, chart editing, line2image, and hybrid.
To ensure semantic diversity within each task, we embed $(I_{in}, Inst)$ into a joint space using Qwen3-VL-Embedding~\citep{qwen3vlembedding} and apply a K-center greedy selection strategy~\citep{sener2018active} to choose $N$ representative samples per task.

\begin{wrapfigure}{r}{0.4\textwidth}
\vspace{-1em}
\centering
\includegraphics[width=\linewidth]{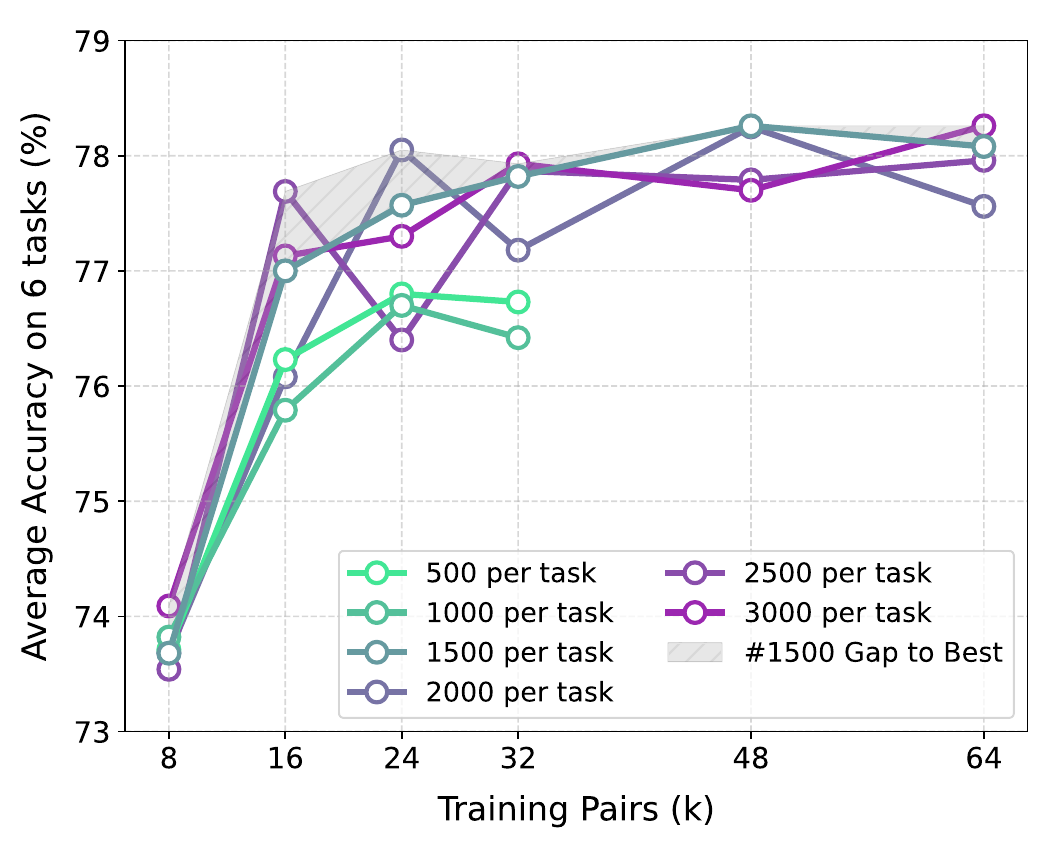}
\caption{\textbf{Average accuracy of different image-instruction pairs per task} on six representative tasks for visual consistency. Performance improves steadily and saturates around 1,500.}
\label{fig:prompt_ablation}
\vspace{-1em}
\end{wrapfigure}
\noindent\textbf{Ablation Study of Pairs Number $N$ per Task.}
Since a small $N$ limits generalization, while a large $N$ substantially increases the cost of downstream preference construction and model training. To identify the most efficient scale, we conduct a targeted ablation study across six representative editing tasks of varying difficulty: subject addition, subject removal, subject replacement, background change, style transfer, and tone transfer.
We scale $N$ from 500 to 3,000 in increments of 500.
For each sampled pair, we randomly choose one generated candidate and directly construct preference pairs using Gemini 3 Pro~\citep{team2023gemini} with the VC pairwise prompt.
Results on EditReward-Bench~\citep{luo2025editscore} in Fig.~\ref{fig:prompt_ablation} show that performance improves steadily up to $N$=1,500 and saturates thereafter, and we therefore set $N$ to 1,500 as a practical trade-off between coverage and efficiency.

In the second stage, for each $(I_{in}, Inst)$, we generate edited outputs using 7 distinct editing models, including BAGEL~\citep{deng2025bagel}, Kontext~\citep{labs2025kontext}, two variants of Step1X-Edit1.2 (preview and standard)~\citep{yin2025reasonedit}, and the Qwen-Image-Edit series (Base, 2509, and 2511)~\citep{wu2025qwenimagetechnicalreport}.
Finally, we constructed $\sim$180k output images as the candidate pool for the following preference data construction.

\subsection{Preference Data Construction Protocol for Visual Consistency}\label{subsec:preference_data}
Our protocol constructs preference pairs across three specialized pipelines: object- and human-centric pipelines for local editing, and a VLM-as-a-Judge approach for global tasks.

\begin{figure}[tb]
    \centering
    \includegraphics[width=1\linewidth]{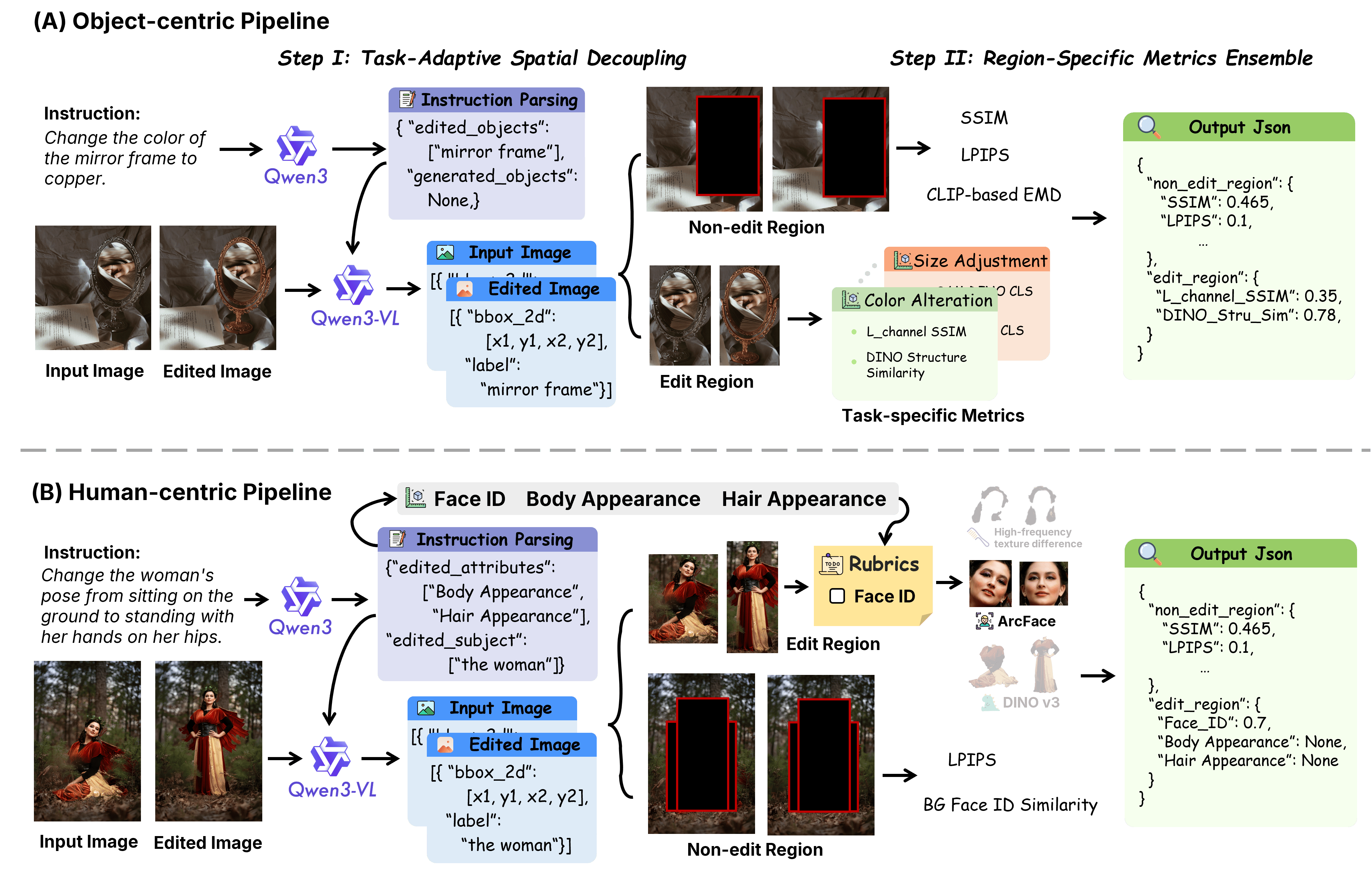}
    \caption{\textbf{Preference data construction pipelines.}
    (A) Object-centric pipeline: instructions are parsed to localize edited entities, partitioning the image into edit and non-edit regions. Region-specific metrics are then applied to enforce background fidelity while evaluating identity consistency within edited areas.
    (B) Human-centric pipeline: extends the object-centric pipeline by decomposing human attributes (face identity, body appearance, hair appearance) and dynamically excluding the edited attribute, enabling fine-grained consistency evaluation using specialized expert models.
    }
    \label{fig:pipeline}
\end{figure}

\noindent\textbf{Object-centric Pipeline.}
As shown in Fig.~\ref{fig:pipeline}(A), this pipeline evaluates instance identity for subject-level tasks (e.g., addition, removal, and attribute modification) through the following two steps.

\textit{$\bullet$ Step I: Task-Adaptive Region Decoupling.}
We first employ Qwen3-4B-Instruct-2507~\citep{qwen3technicalreport} to extract the editing target from the instruction.
Given that different editing operations affect the input image and output image asymmetrically, we utilize Qwen3-VL-8B-Instruct~\citep{qwen3_vl} to perform task-adaptive grounding.
For example, the editing target is localized in the input image for removal, in the output image for addition, and in both for replacement or attribute modification.
This process partitions the image into an `Edit Region' ($\Omega_{edit}$), encompassing the union of the localized masks and a `Non-edit Region' ($\Omega_{non}$), representing the remaining background.
This decoupling ensures that our evaluation is spatially anchored to the areas where consistency is most critical.

\textit{$\bullet$ Step II: Region-Specific Metrics Ensemble.}
Building on the region partition, we apply a dual-strategy metric to assess consistency without penalizing the intended edit.
In $\Omega_{non}$, we enforce strict visual invariance using a combination of SSIM~\citep{wang2004image}, LPIPS~\citep{zhang2018unreasonable}, and CLIP-based Earth Mover's Distance~\citep{rubner1998metric} (EMD) to ensure both low-level visual features and high-level semantic content remain unchanged.
Conversely, within $\Omega_{edit}$, we utilize task-specific metrics to decouple identity preservation from the editing effects; for instance, in the color alteration task, we compute SSIM exclusively on the lightness channel to assess structural integrity while allowing for the chromatic shifts required by the instruction.
The mapping of these task-to-metric configurations is detailed in the Appendix~\ref{subsec:pipeline}.

\noindent\textbf{Human-centric Pipeline.}
For tasks involving human subjects, as shown in Fig.~\ref{fig:pipeline}(B), we extend the object-centric pipeline by decomposing human visual properties into three orthogonal attributes: \textit{Face IDentity} (Face ID), \textit{Body Appearance}, and \textit{Hair Appearance}.
While inheriting the spatial decoupling logic described above, this specialization requires a more granular instruction-parsing phase: the language model identifies not only the target subjects but also the specific attribute slated for modification.
This allows us to derive a dynamic, attribute-conditional rubric within $\Omega_{edit}$ by automatically excluding the modified property from the three attributes.
The remaining stationary attributes are then quantified using specialized expert models, such as ArcFace~\citep{deng2019arcface} for Face ID and selfie segmenter for Body Appearance.
While $\Omega_{non}$ remains governed by the similarity metrics defined above, this specialization generates reliable preference supervision that isolates intended human edits from unintended identity bleeding or anatomical distortions.

\begin{figure}[t]
    \centering
    \includegraphics[width=1\linewidth]{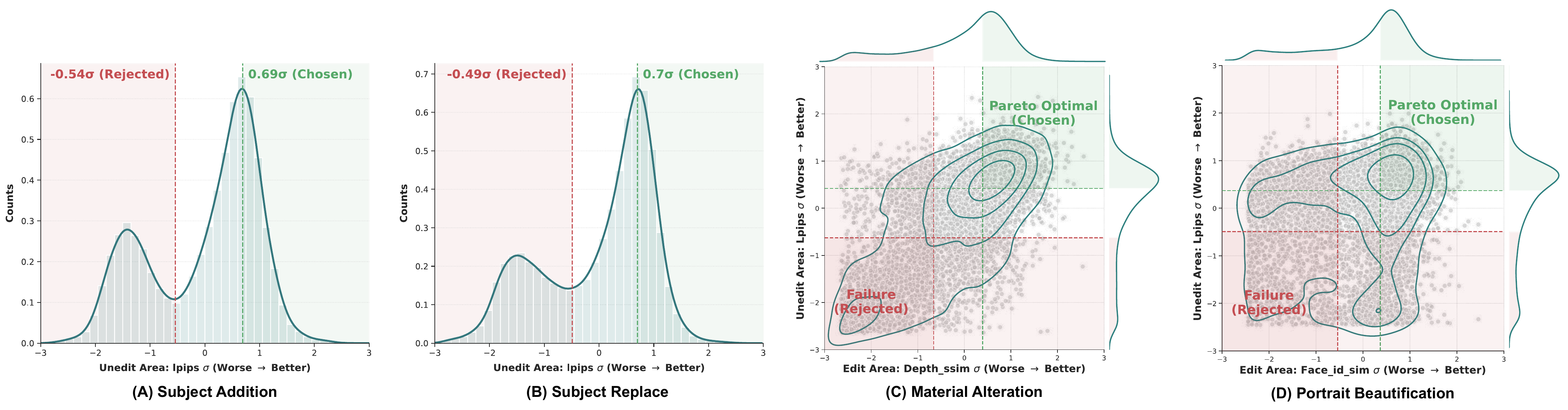}
    \caption{\textbf{Task-wise primary score distributions for preference pair synthesis.} (a-b) Shaded regions indicate the top/bottom 30\% Z-score threshold used to identify Winners (green) and Losers (red) with a clear optimization margin for preference learning. (c-d) Combined with Pareto filtering, candidate pairs that only exhibit score separation across two primary metrics are retained.}
    \label{fig:pairs_distribution}
\end{figure}

\noindent\textbf{Preference Pair Synthesis.}
To transform raw scores into high-confidence preference pairs, we implement a synthesis approach based on statistical distribution and multi-metric consensus.
For each task, we empirically select one primary metric in $\Omega_{non}$ and $\Omega_{edit}$, respectively (e.g., LPIPS for $\Omega_{non}$ and Face ID for human $\Omega_{edit}$), while treating all other metrics as auxiliary validators.
The construction process begins with task-wise z-score normalization of primary metrics within each $(I_{in}, Inst)$ group to mitigate inter-group variance.
These scores are then aggregated with candidates in the top and bottom 30\% of the resulting distribution identified as Winners and Losers, respectively.
The empirical score distributions for four representative tasks, as visualized in Fig.~\ref{fig:pairs_distribution}, further validate the efficacy of this thresholding in capturing clear optimization margins.
To reconcile potentially conflicting regional requirements, we enforce a strict Pareto Dominance rule: a candidate pair $(I_A, I_B)$ is used only if $I_A$ is superior to $I_B$ in at least one regional primary metric without being inferior in the other.
Finally, the derived preference is cross-validated through majority voting across all auxiliary validators; any pair exhibiting a conflict between primary and auxiliary indicators is discarded.
Such dual-layer filtering ensures that the resulting preference pairs exhibit a significant optimization margin and are grounded in consistent visual evidence across all evaluated dimensions.

\noindent\textbf{VLM-as-a-Judge Annotation.}
For global editing tasks where localized masking and region-specific metrics are inapplicable, we shift from bottom-up pixel-wise measures to top-down semantic reasoning.
Specifically, for each candidate pair, we use Gemini 3 Pro to perform pairwise consistency assessments.
However, for each $(I_{in}, Inst)$, evaluating the full set of $\binom{7}{2}$ possible candidate pairs
\begin{wrapfigure}{r}{0.4\textwidth}
\centering
\vspace{-1em}
\includegraphics[width=\linewidth]{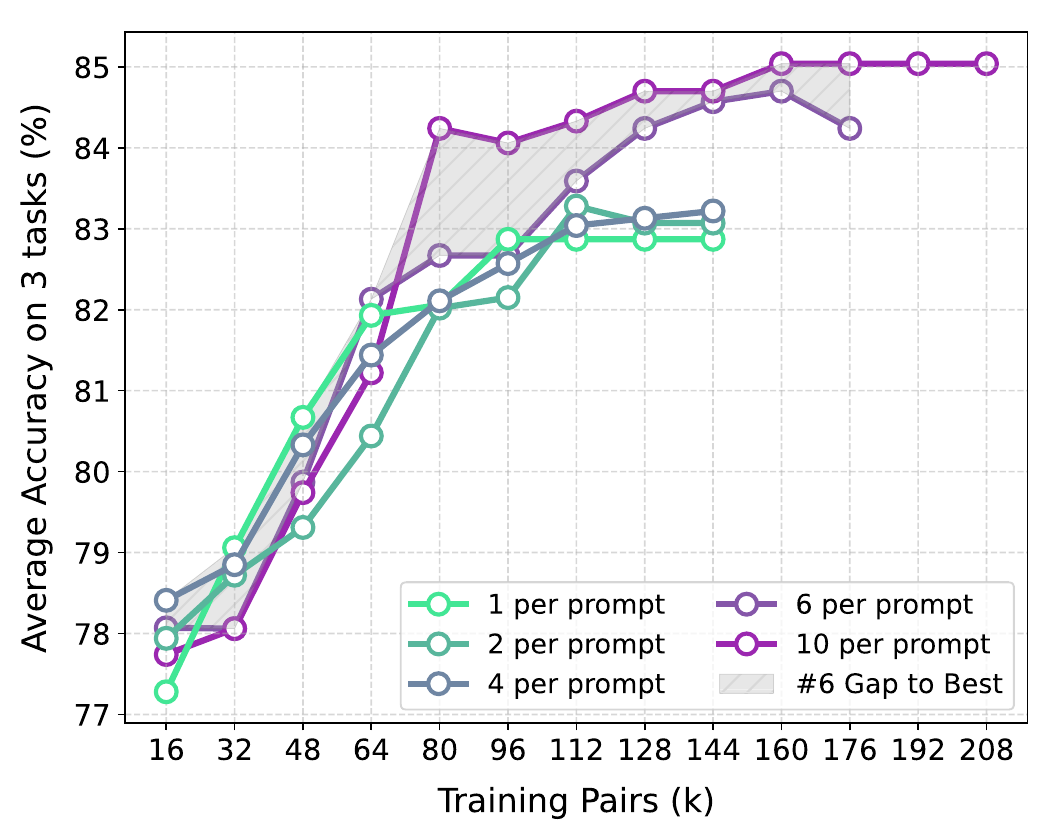}
\caption{\textbf{Average accuracy on three tasks of different in-group pairs ($P$)} for visual consistency. $P$=6, chosen as the optimal balance between supervision density and computation cost.}
\label{fig:pair_ablation}
\vspace{-1em}
\end{wrapfigure}
entails prohibitive annotative costs and significant computational overhead.
To determine the optimal balance between \textit{in-group pair diversity and efficiency}, we conduct an ablation study across three representative global tasks: background change, style transfer, and tone transfer.
Fixing the sample size at $N$ = 1,500, we vary the number of sampled in-group pairs $P$ within $\{1, 2, 4, 6, 10\}$.
The evolution of average accuracy on EditReward-Bench~\citep{luo2025editscore} relative to training steps is shown in Fig.~\ref{fig:pair_ablation}.
We find that performance remains stagnant for $P \le$ 4 but exhibits a significant leap at $P$ = 6, beyond which marginal gains become negligible.
We attribute this improvement to the enriched intra-group diversity, which potentially provides more fine-grained preference signals conducive to robust discriminative learning.
Thus, we fix $P$ to 6 as the optimal trade-off between supervisorial density and computational cost.
Finally, we constructed $\sim$128k preference pairs for training.

\subsection{Training Details}\label{subsec:training}
Our final PVC-Judge is obtained by fine-tuning Qwen3-VL-8B-Instruct~\citep{qwen3_vl} with LoRA~\citep{hu2022lora} on our curation preference dataset described in Sec.~\ref{subsec:preference_data}.
We adopt the AdamW optimizer with a learning rate of $2.0 \times 10^{-6}$, together with a cosine learning rate scheduler and a warmup ratio of 0.05.
The model is trained for 3 epochs using an effective batch size of 16, achieved via a per-device batch size of 2  across 8 NVIDIA L40S GPUs.
For parameter-efficient adaptation, we set the LoRA rank ($r$) to 64, balancing adaptation capacity and training efficiency.

\subsection{VCReward-Bench}\label{subsec:vcreward_bench}

\begin{wraptable}{r}{5cm}
    \centering
    \vspace{-1.5em}
    \caption{\textbf{Comparison of VCReward-Bench with existing reward benchmarks.}
    }
    \label{tab:reward_benchmark_comparison}
    \resizebox{\linewidth}{!}{
    \begin{tabular}{lcc}
        \toprule
        \textbf{Benchmark} & \textbf{Size} & \textbf{Tasks} \\
        \midrule
        EditScore~\citep{luo2025editscore} & 3K & 11  \\
        EditReward~\citep{wu2025editreward} & 1.5K & -  \\
        \textbf{VCReward-Bench} & \textbf{3.5K} & \textbf{21} \\
        \bottomrule
    \end{tabular}
    }
    \vspace{-1em}
\end{wraptable}
To evaluate the alignment of PVC-Judge with human judgment, we build VCReward-Bench, a high-quality meta-evaluation set that offers a more comprehensive taxonomy and larger scale than previous benchmarks like EditReward-Bench in~\citep{luo2025editscore}~and~\citep{wu2025editreward} (see Table~\ref{tab:reward_benchmark_comparison}).
Adopting the 21 predefined tasks in Sec.~\ref{subsec:taxonomy}, we collect image-instruction pairs by repurposing existing image editing benchmarks for most categories.
For newly defined tasks such as chart editing, we harvest raw images from the web and employ trained experts to craft precise instructions.
Candidate images for each input pair are synthesized using the 7 editing models used in Sec.~\ref{subsec:image_gen} alongside Nano Banana Pro~\citep{team2023gemini}, ensuring a broad spectrum of editing artifacts.
Unlike the automated protocol used for training data, VCReward-Bench is entirely annotated by human experts through rigorous pairwise comparisons (the detailed annotation process is provided in the Appendix~\ref{sec:vcReward_bench}).
Finally, VCReward-Bench contains 3,506 preference pairs across 21 tasks, serving as a robust testbed for evaluating the assessment models of image editing in visual consistency.

\section{Experiments}
In this section, we first conduct a meta-evaluation of human agreement for our PVC-Judge and report the leaderboard on GEditBench v2 across three evaluation dimensions.
Additional qualitative analyses of challenging applications, including \textbf{open-set edits}, \textbf{spatial relation perception}, and \textbf{small-face consistency}, are provided in Appendix~\ref{subsec:add_qualitative}.

\begin{figure}[tb]
    \centering
    \includegraphics[width=1\linewidth]{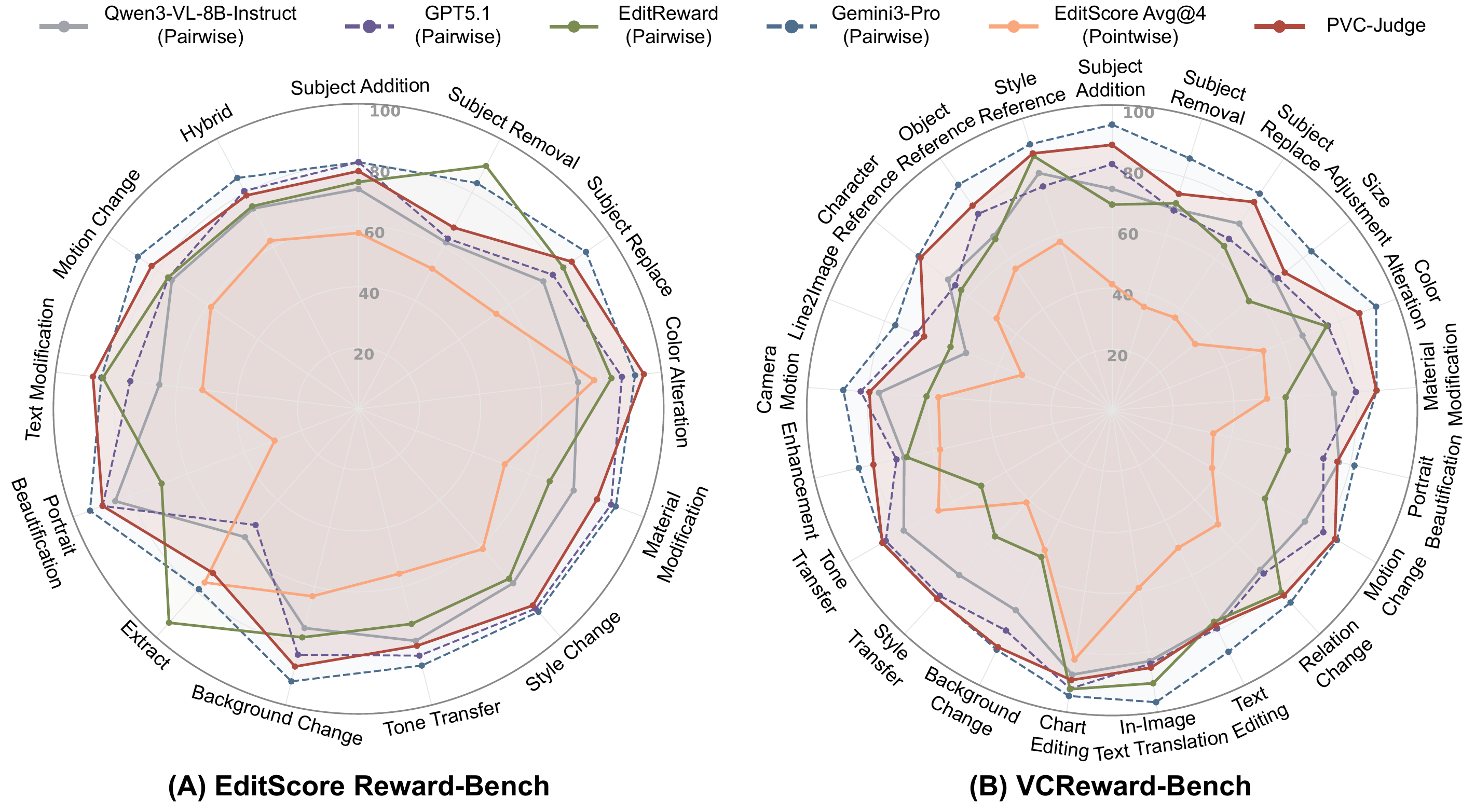}
    \caption{\textbf{Human alignment of assessment models for visual consistency} on (a) EditReward-Bench~\citep{luo2025editscore} and (b) VCReward-Bench. Notably, our model achieves state-of-the-art performance among open-source evaluators across nearly all tasks, performing on par with or even surpassing the proprietary GPT-5.1-2026-02-28.}
    \label{fig:reward_res}
\end{figure}

\subsection{Meta-Evaluation of PVC-Judge}
\noindent{\textbf{Setup.}}
To evaluate the effectiveness of our PVC-Judge, we compare it against Qwen-3-VL-8B-Instruct, GPT-5.1, Gemini 3 Pro, and two editing reward models, i.e., the EditScore~\citep{luo2025editscore} (Qwen-3-VL-8B version) and EditReward~\citep{wu2025editreward} (MiMo-VL-7B-SFT-2508 version).
For a fair comparison, the three VLMs employ the exact same pairwise comparison prompt as PVC-Judge.
For EditScore and EditReward, their original author-provided prompts are adopted.
Besides, we further enhance EditScore with its recommended self-ensembling strategy to establish a better baseline, denoted as EditScore-Avg@4.
To assess human alignment, we evaluate the models on our VCReward-Bench and the visual consistency subset of EditReward-Bench~\citep{luo2025editscore}.

\noindent{\textbf{Results.}}
Comparison results in Fig.~\ref{fig:reward_res} reveal three key observations:
1) PVC-Judge consistently outperforms its base model, Qwen3-VL-8B-Instruct, across all tasks on both benchmarks, validating the quality of our synthesized preference pairs and data curation pipeline.
2) Compared with existing reward models, our model achieves higher accuracy on most tasks, indicating that PVC-Judge provides more precise and human-aligned evaluation for visual consistency.
3) Despite its modest 8B size, our model achieves performance on par with leading proprietary VLMs and even surpasses GPT-5.1 on average. Appendix~\ref{subsec:add_numerial} provides the detailed numerical results.

\subsection{Main Results on GEditBench v2}
\noindent\textbf{Setup.}
We evaluate 16 representative image editing models on \textbf{GEditBench v2} to provide a comprehensive evaluation of current capabilities.
These models include three leading closed-source models, GPT-Image-1.5~\citep{gpt4o20250325}, Nano Banana Pro~\citep{team2023gemini}, and Seedream4.5~\citep{seedream2025seedream}, as well as thirteen open-source models: BAGEL~\citep{deng2025bagel}, OmniGen2~\citep{wu2025omnigen2}, Kontext~\citep{labs2025kontext}, Step1x-Edit-v1p2~\citep{yin2025reasonedit}, three versions of Qwen-Image-Edit (Base, 2509, 2511)~\citep{wu2025qwenimagetechnicalreport}, FLUX.2 [dev]~\citep{flux2_dev}, Flux.2 [dev] Turbo, LongCat-Image-Edit~\citep{team2025longcat_image}, GLM-Image~\citep{GLM_Image}, along with two efficiency-oriented step-distilled variants, FLUX.2 [klein] 4B/9B~\citep{flux2_klein}.
Default hyperparameter settings are used to ensure fairness.

For ranking methodology, rather than relying on simple win-rates, following~\citep{zheng2023judging}, we first utilize the Bradley-Terry (BT) model~\citep{bradley1952rank} to estimate the underlying capability score for each editing model based on the aggregated win/tie/loss matrix.
These capability scores are then transformed into standard Elo ratings to provide an intuitive and globally comparable measure of relative performance.
Furthermore, to account for statistical variance and sampling noise, we compute the 95\% Confidence Intervals (CI) for each model's ELO rating via 1,000 bootstrapping iterations over the evaluation samples, ensuring that our final rankings reflect statistically significant performance differences.
We construct the Overall metric by fitting a shared BT model on the aggregated pairwise comparisons across all evaluation dimensions.

\noindent\textbf{Leaderboard.}
\begin{table}[tb]
    \centering
    \caption{\textbf{OpenEdit leaderboard.} Models are ranked by Elo ratings derived from a pairwise comparison paradigm. Instruction Following and Visual Quality are assessed by GPT-4o (26-03-24), while Visual Consistency is exclusively evaluated by our proposed PVC-Judge. Overall Elo scores and their 95\% Confidence Intervals (CI) are computed via 1,000 bootstrap iterations. \close means closed-source, \open means open-source. $^\star$The Elo scores of Arena were recorded on March 26, 2026.}
    \label{tab:leadboard}
    \resizebox{\linewidth}{!}{
    \begin{tabular}{lccccccc>{\columncolor{gray!20}}c>{\columncolor{gray!20}}ccc}
    \toprule
     \multirow{2}{*}{\textbf{Model}} & \multirow{2}{*}{\textbf{Samples}} & \multicolumn{2}{c}{\textbf{Instruction Following}} & \multicolumn{2}{c}{\textbf{Visual Quality}} & \multicolumn{2}{c}{\textbf{Visual Consistency}} & \multicolumn{2}{c}{\cellcolor{gray!20}\textbf{Overall}} & \multicolumn{2}{c}{\textbf{Arena$^\star$}}\\
     & & ELO$\uparrow$ & 95\% CI & ELO$\uparrow$ & 95\% CI & ELO$\uparrow$ & 95\% CI & ELO$\uparrow$ & 95\% CI & ELO$\uparrow$ & Rank \\
    \midrule
    \close Nano Banana Pro (26-03-04) & 1,156 & 1,126 & -13/+15 & 1,066 & -9/+10 & 1,108 & -11/+11 & 1,096 & -6/+6 & 1,251 & 2 \\
    \close Seedream 4.5 (26-03-11) & 1,190 & 1,111 & -12/+12 & 1,142 & -11/+11 & 1,030 & -11/+12 & 1,089 & -7/+7 & 1,196 & 3 \\
    \close GPT Image 1.5 (26-03-04) & 1,081 & 1,260 & -13/+15 & 1,149 & -12/+12 & 846 & -13/+13 & 1,071 & -7/+6 & 1,270 & 1 \\
    \open FLUX.2 [klein] 9B & 1,200 & 1,083 & -13/+12 & 1,025 & -11/+10 & 1,019 & -10/+9 & 1,039 & -6/+6 & 1,166 & 4 \\
    \open Qwen-Image-Edit-2511 & 1,200 & 1,095 & -10/+10 & 1,060 & -11/+11 & 972 & -9/+10 & 1,038 & -6/+6 & 1,164 & 5\\
    \open FLUX.2 [klein] 4B & 1,200 & 1,007 & -12/+12 & 1,019 & -10/+10 & 1,070 & -10/+10 & 1,031 & -6/+6 & 1,107 & 10\\
    \open FLUX.2 [dev] Turbo & 1,200 & 1,068 & -12/+12 & 936 & -10/+10 & 1,064 & -11/+10 & 1,021 & -6/+6 & 1,153 & 6\\
    \open Qwen-Image-Edit-2509 & 1,200 & 1,033 & -10/+11 & 1,062 & -10/+12 & 955 & -9/+9 & 1,014 & -5/+6 & 1,142 & 7\\
    \open Qwen-Image-Edit & 1,200 & 991 & -10/+10 & 1,073 & -11/+12 & 971 & -11/+11 & 1,010 & -6/+6 & 1,088 & 12\\
    \open FLUX.2 [dev] & 1,200 & 1,037 & -12/+13 & 965 & -10/+10 & 1,018 & -11/+11 & 1,006 & -7/+7 & 1,137 & 8\\
    \open LongCat-Image-Edit & 1,200 & 1,018 & -10/+11 & 968 & -10/+9 & 1,017 & -10/+9 & 1,001 & -6/+5 & 1,111 & 9\\
    \open Step1X-Edit-v1p2 & 1,200 & 909 & -12/+12 & 1,007 & -12/+11 & 1,067 & -11/+11 & 996 & -6/+7 & 1,093 & 11\\
    \open GLM-Image & 1,200 & 787 & -13/+14 & 1,023 & -11/+11 & 1,109 & -13/+14 & 979 & -6/+6 & 930 & 14\\
    \open OmniGen V2 & 1,200 & 807 & -13/+12 & 910 & -12/+12 & 929 & -13/+13 & 888 & -7/+7 & 919 & 15\\
    \open FLUX.1 Kontext [dev] & 1,200 & 849 & -13/+13 & 900 & -13/+14 & 840 & -14/+13 & 869 & -7/+8 & 1,017 & 13\\
    \open Bagel & 1,200 & 820 & -13/+13 & 694 & -17/+16 & 987 & -13/+14 & 851 & -8/+8 & 915 & 16\\
    \bottomrule
    \end{tabular}
    }
\end{table}
Table~\ref{tab:leadboard} presents the final evaluation results ranked by the Overall ELO score, including IF and VQ ELO scores obtained via GPT-4o pairwise comparisons, as well as VC ELO scores evaluated by our PVC-Judge.
For a broader context, we include human-annotated Arena~\citep{arena} Elo scores as a reference.
Although Arena scores derive from different test sets, our Overall Elo achieves a strong Spearman's rank correlation ($\rho$=0.929, $p{<}$2e-7) with the Arena rankings, validating that our automated evaluation ecosystem reliably aligns with human preferences.

At the apex of the rankings, proprietary models continue to set the performance ceiling. Nano Banana Pro secures the definitive first place, closely followed by Seedream 4.5 with only a marginal capability gap.
Strikingly, the open-source community demonstrates remarkable competitiveness through architectural efficiency. FLUX.2 [klein] 9B, a 4-step distilled model, emerges as the open-source champion and successfully narrows the gap with proprietary giants.
It maintains a narrow lead over formidable open-source alternatives like Qwen-Image-Edit-2511, highlighting the immense potential of lightweight models for high-quality image editing.

Deconstructing the Overall Elo into its constituent dimensions reveals the inherent trade-off between aggressive instruction execution and visual consistency.
Models like GLM-Image and Bagel perfectly illustrate an ``under-editing'' trap.
They achieve artificially inflated VC scores (1,109 and 987, respectively) precisely because their diminished IF capabilities (787 and 820) prevent them from meaningfully modifying the input image.
This phenomenon strongly validates the necessity of this multi-dimensional ranking ecosystem.
It suggests that evaluating visual consistency in isolation remains insufficient for assessing true editing proficiency.
Conversely, top-tier models like Nano Banana Pro and the distilled FLUX.2 [klein] variants successfully navigate this complex trade-off.
They secure top overall rankings by maintaining a delicate equilibrium between executing complex user prompts and preserving unintended elements seamlessly.
\section{Conclusion}
In this paper, we present a unified evaluation ecosystem to address the evaluative mismatch in instruction-based image editing.
We introduce \textbf{GEditBench v2}, a 23-task benchmark that moves beyond constrained settings toward complex, open-set real-world scenarios.
To enable reliable assessment for visual consistency, we propose \textbf{PVC-Judge}, an open-source pairwise evaluation model.
Powered by two novel region-decoupled preference data synthesis pipelines, our model achieves strong human alignment in visual consistency.
We further establish \textbf{VCReward-Bench}, designed to evaluate assessment models of image editing in visual consistency.
In future work, we plan to integrate PVC-Judge into the training loop as a reward model for precise image editing.

\bibliography{iclr2026_conference}
\bibliographystyle{iclr2026_conference}

\newpage
\appendix
\section*{Appendix Overview}

In this supplemental material, we provide additional details omitted from the main text:

\appsection{sec:limit}{A. Limitations}

\appsection{sec:openedit}{B. GEditBench v2}
\appsubsection{subsec:task_def}{B.1. Detailed Tasks Explanation}
\appsubsection{subsec:prompts}{B.2. Pairwise Evaluation Prompts}

\appsection{sec:pvc_judge}{C. Pairwise Visual Consistency Judge}
\appsubsection{subsec:pipeline}{C.1. \texttt{(task, Pipeline, Region-Specific Metrics)} Mapping}
\appsubsection{subsec:data_distri}{C.2. Preference Data Distribution}
\appsubsection{subsec:train_hyper}{C.3. Training Hyper-parameters}

\appsection{sec:vcReward_bench}{D. Annotation Protocol for VCReward-Bench}

\appsection{sec:add_results}{E. Additional Results}
\appsubsection{subsec:add_numerial}{E.1. Full Numerical Results of Meta-Evaluation}
\appsubsection{subsec:add_qualitative}{E.2. Qualitative Analysis}

\newpage

\section{Limitations}\label{sec:limit}
While GEditBench v2 establishes a general image editing benchmark built on real-world instructions and develops an assessment model for visual consistency, it may still exhibit several limitations.

First, evaluating large-scale image editing models requires substantial computational resources and long inference times.
To maintain practicality, we limit the current benchmark size, which may reduce the diversity of test samples within individual editing tasks.
Future work will expand the dataset while exploring more efficient evaluation and sampling strategies.

Second, our automated object- and human-centric pipelines for constructing preference pairs rely on several pre-trained foundation models, e.g., SAM, CLIP, and DINOv3, to extract regional features.
While these models enable scalable data construction, they may also introduce potential biases into the resulting preference dataset.
Mitigating such inherited biases and improving the robustness of the underlying feature extractors will be an important direction for future iterations.

\section{GEditBench v2}\label{sec:openedit}

\subsection{Detailed Tasks Explanation}\label{subsec:task_def}
GEditBench v2 establishes a comprehensive evaluation taxonomy comprising 23 distinct editing tasks.
We distribute these tasks across five fundamental categories: (1) Local, (2) Global, (3) Reference, (4) Hybrid, and a new (5) Open-set category.
This multi-level design strictly challenges both basic manipulation edits and advanced real-world instruction understanding.
Below, we comprehensively explain each task.

\begin{figure}[tb]
    \centering
    \includegraphics[width=1\linewidth]{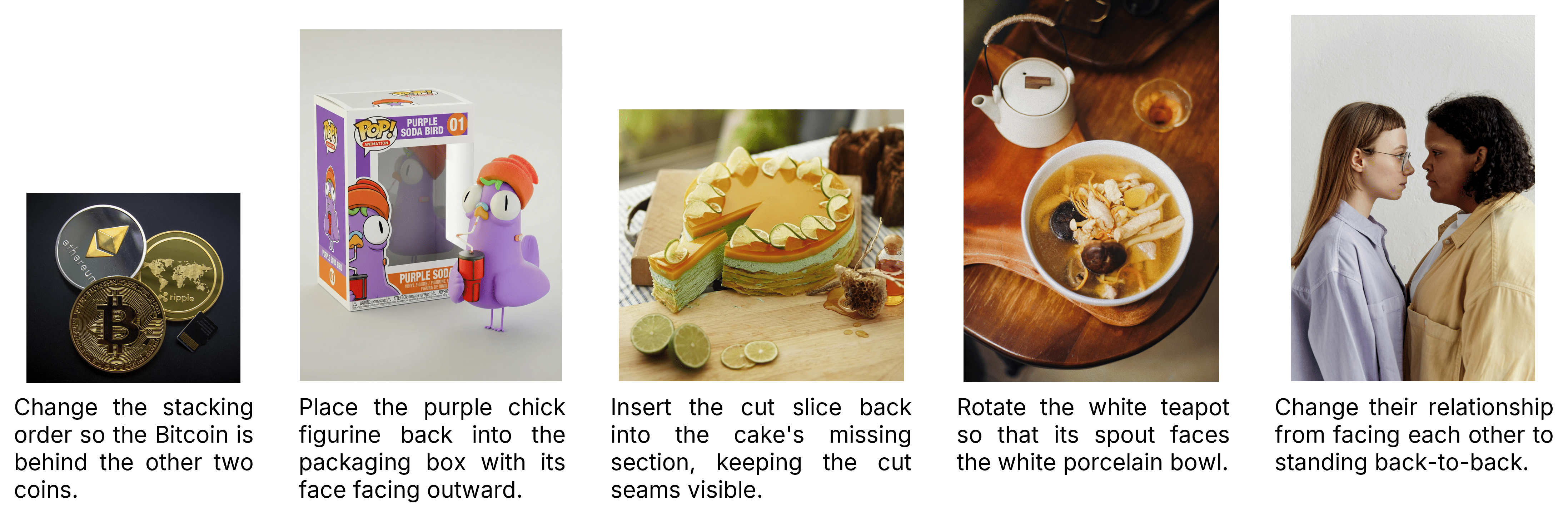}
    \caption{\texttt{(Input Image, Instruction)} examples for the \textit{Relation Change} task.}
    \label{fig:relation_case}
\end{figure}

\begin{figure}[tb]
    \centering
    \includegraphics[width=1\linewidth]{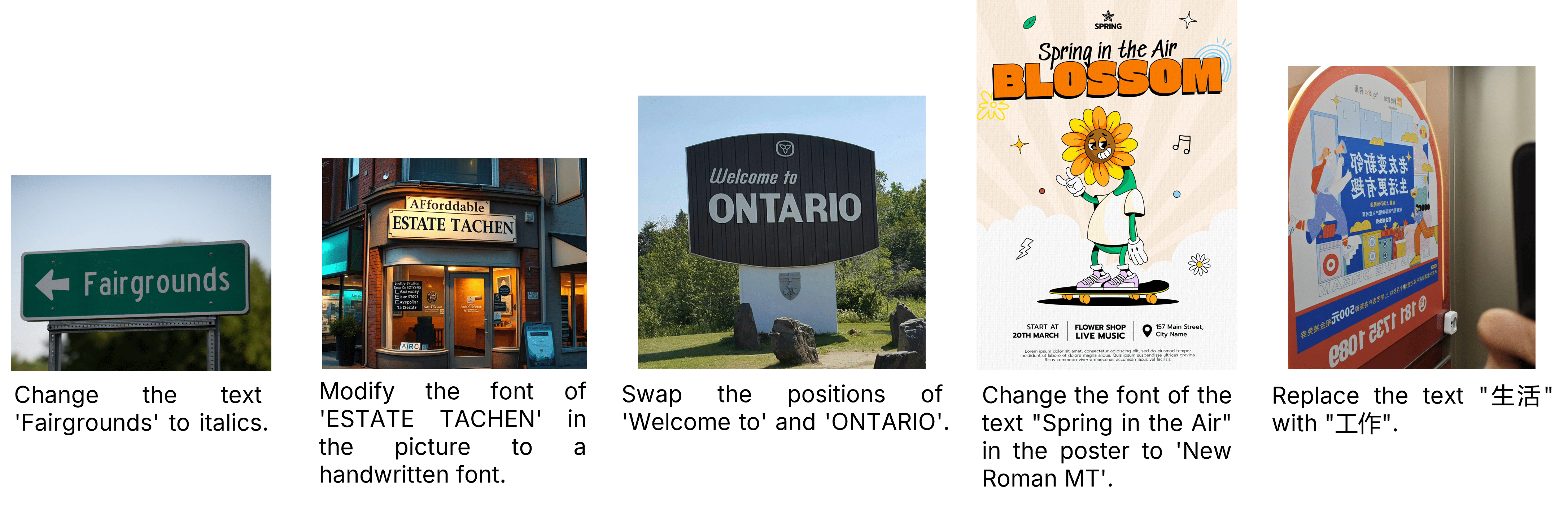}
    \caption{\texttt{(Input Image, Instruction)} examples for the \textit{Text Editing} task. Our benchmark comprehensively evaluates fundamental semantic modifications alongside highly demanding typographic layout changes and precise font alterations.}
    \label{fig:text_editing_case}
\end{figure}

\begin{figure}[!t]
    \centering
    \includegraphics[width=1\linewidth]{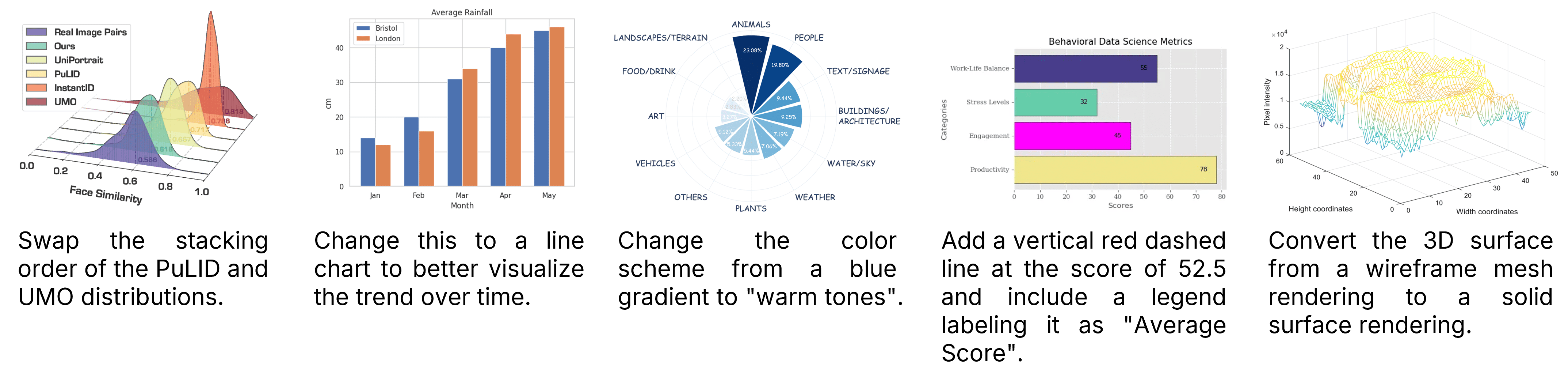}
    \caption{\texttt{(Input Image, Instruction)} examples for the \textit{Chart Editing} task.}
    \label{fig:chart_editing_case}
\end{figure}

\noindent\textbf{Local Editing.}
This category evaluates spatially restricted modifications that target specific image regions or objects across the following 12 tasks.
\begin{enumerate}
    \item \textit{Subject Addition}: Seamlessly inserts a newly designated entity into a specific location within the original scene.
    \item \textit{Subject Removal}: Erases a specified target object and plausibly reconstructs the newly exposed underlying region.
    \item \textit{Subject Replace}: Swaps an existing object with a completely new target entity while strictly preserving the original spatial layout.
    \item \textit{Size Adjustment}: Scales a specific object up or down without distorting its inherent structural integrity.
    \item \textit{Color Alteration}: Modifies the hue of a targeted region while retaining its original physical texture and lighting.
    \item \textit{Material Modification}: Transforms the surface properties of a specific object to reflect a completely different physical material.
    \item \textit{Portrait Beautification}: Applies targeted aesthetic enhancements to human subjects without altering their underlying recognizable identity.
    \item \textit{Motion Change}: Alters the physical posture or dynamic action of a specified subject within the visual environment.
    \item \textit{Relation Change}: Modifies the spatial positioning or physical interaction dynamics between multiple existing entities.
    More examples are provided in Fig.~\ref{fig:relation_case}.
    \item \textit{Text Editing}: Executes the fundamental addition, deletion, or semantic modification of text elements alongside targeted changes to their typographic styles and spatial layouts. Some cases are shown in Fig.~\ref{fig:text_editing_case}.
    \item \textit{In-Image Text Translation}: Accurately translates embedded textual elements into a target language while seamlessly preserving the original typographic aesthetics and background context.
    \item \textit{Chart Editing}: Transforms data visualizations through targeted aesthetic enhancements or structural chart type conversions without compromising the underlying informational integrity (e.g., numerical values and geometric proportions).
    Fig.~\ref{fig:chart_editing_case} visualizes representative examples.
\end{enumerate}

\begin{figure}[thb]
    \centering
    \includegraphics[width=1\linewidth]{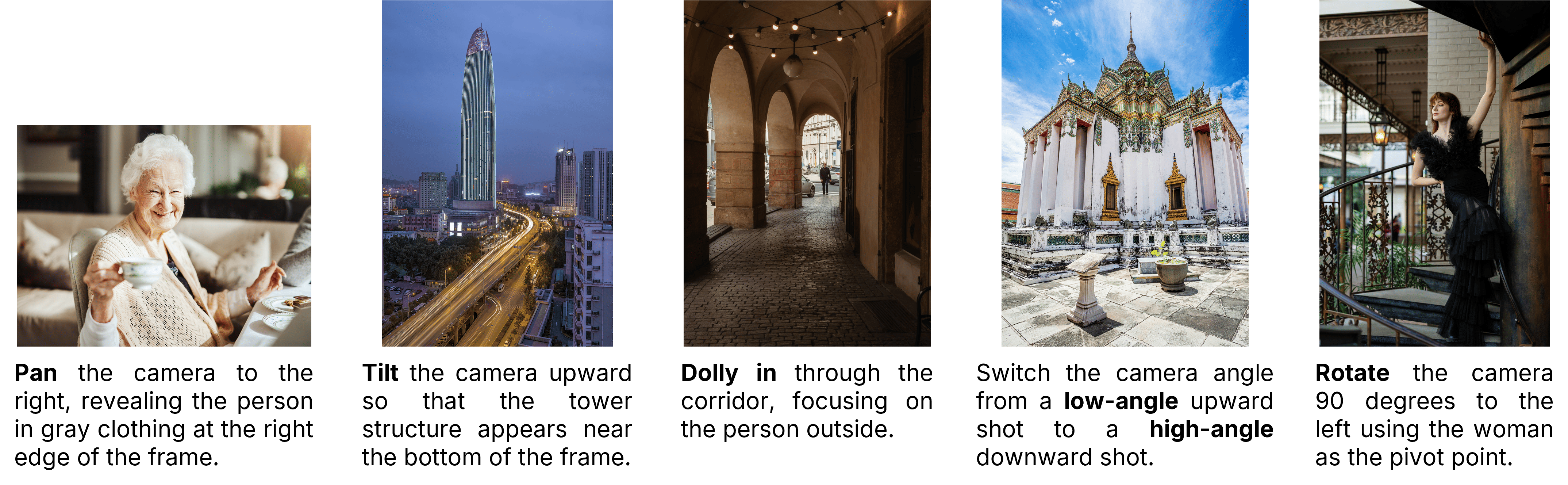}
    \caption{\texttt{(Input Image, Instruction)} examples for the \textit{Camera Motion} task.}
    \label{fig:camera_motion}
\end{figure}

\begin{figure}[t]
    \centering
    \includegraphics[width=1\linewidth]{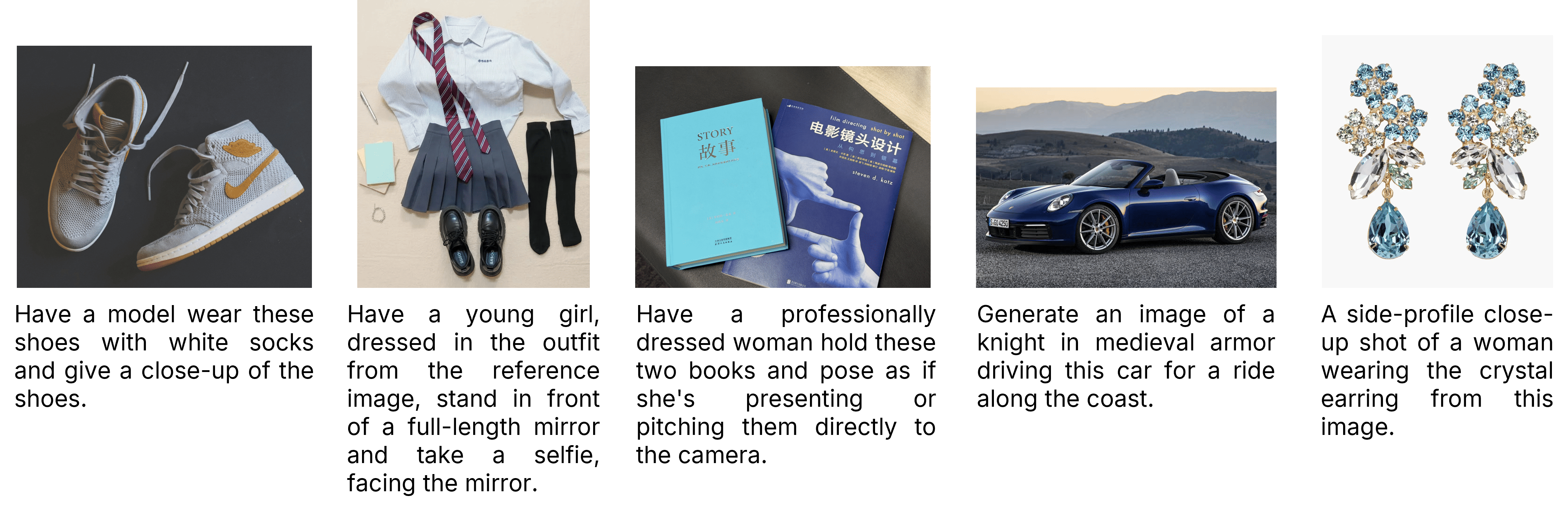}
    \caption{\texttt{(Input Image, Instruction)} examples for the \textit{Object Reference} task.}
    \label{fig:oref_cases}
\end{figure}

\begin{figure}[t]
    \centering
    \includegraphics[width=1\linewidth]{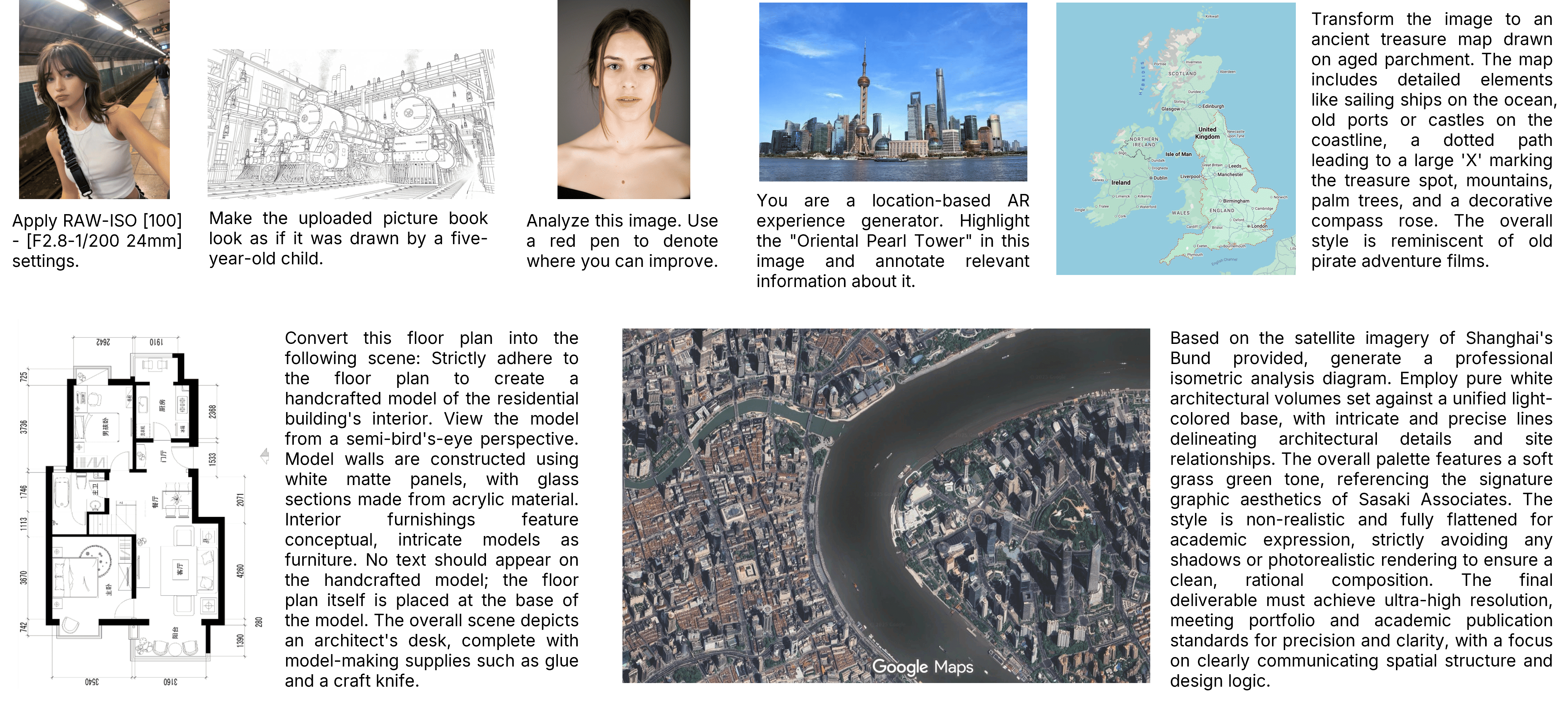}
    \caption{\texttt{(Input Image, Instruction)} examples for the \textit{Open-Set} task.}
    \label{fig:openset_cases}
\end{figure}

\noindent\textbf{Global Editing.}
This editing paradigm evaluates holistic transformations that alter the entire visual atmosphere or structural layout of the image across the following 6 tasks.
\begin{enumerate}[start=13]
    \item \textit{Background Change}: Replaces the contextual environment surrounding the main subjects with diverse natural and cultural settings across both indoor and outdoor scenarios.
    \item \textit{Style Transfer}: Imposes 17 distinct artistic and aesthetic styles across the entire image while rigorously maintaining the underlying geometric layout.
    \item \textit{Tone Transfer}: Shifts the global lighting conditions, color palettes, seasonal atmospheres, and weather dynamics to evoke a completely different environmental mood.
    \item \textit{Enhancement}: Restores visual fidelity by eradicating 9 specific low-level degradations, including blur, compression, moiré, low-light, noise, flare, reflection, haze, and rain alongside complex old photo restoration and rigorous overexposure repair.
    \item \textit{Camera Motion}: Simulates physical camera movements encompassing typical zoom operations alongside complex pan, rotate, tilt, and special view switches (see Fig.~\ref{fig:camera_motion}).
    \item \textit{Line2Image}: Transforms sparse structural sketches or edge maps into fully rendered outputs with coherent global textures and lighting.
\end{enumerate}

\noindent\textbf{Reference Editing.}
This category rigidly evaluates the capacity of editing models to extract and faithfully transfer specific visual identities from an external guiding image across the following 3 tasks.

\begin{enumerate}[start=19]
    \item \textit{Character Reference}: Synthesizes a specific person from a reference image across novel scenes and states while perfectly preserving their exact identity and characteristics from the reference image.
    \item \textit{Object Reference}: Synthesizes a specific object from the reference image across novel scenes and states while perfectly preserving its exact physical details. Fig.~\ref{fig:oref_cases} visualizes representative examples.
    \item \textit{Style Reference}: Applies the visual style of the reference image to generate the target scene without copying its actual objects or content.
\end{enumerate}

\noindent\textbf{Hybrid Editing.}
This task (\textit{Hybrid}) evaluates compositional editing instructions that require multiple operations within a single query.
Each prompt combines $3\sim5$ predefined tasks from earlier categories into one unified instruction.
This setting examines whether models can reliably execute multiple edits on the same image without omitting or conflating distinct semantic constraints.

\noindent\textbf{Open-Set Editing.}
This open-set category is designed to evaluate model behavior under more general and flexible editing instructions that cannot be easily categorized into predefined task types.
While the previous categories cover a broad range of structured editing operations, real-world queries often involve mixed intents or loosely specified objectives.
These instructions do not strictly follow a fixed task taxonomy and, therefore, provide a useful test of model generalization beyond predefined settings.
To construct this set, we manually curated 100 diverse prompts collected from public online sources, including platforms such as X, Reddit, and community-curated GitHub repositories.
These prompts were selected to reflect instruction patterns that are difficult to assign to a single editing category, often combining multiple intents or expressing goals at a higher semantic level (see Fig.~\ref{fig:openset_cases}).
As a result, the open-set subset provides a complementary evaluation setting that better reflects the flexibility required in practical editing scenarios.

\begin{figure}[t]
    \centering
    \includegraphics[width=1\linewidth]{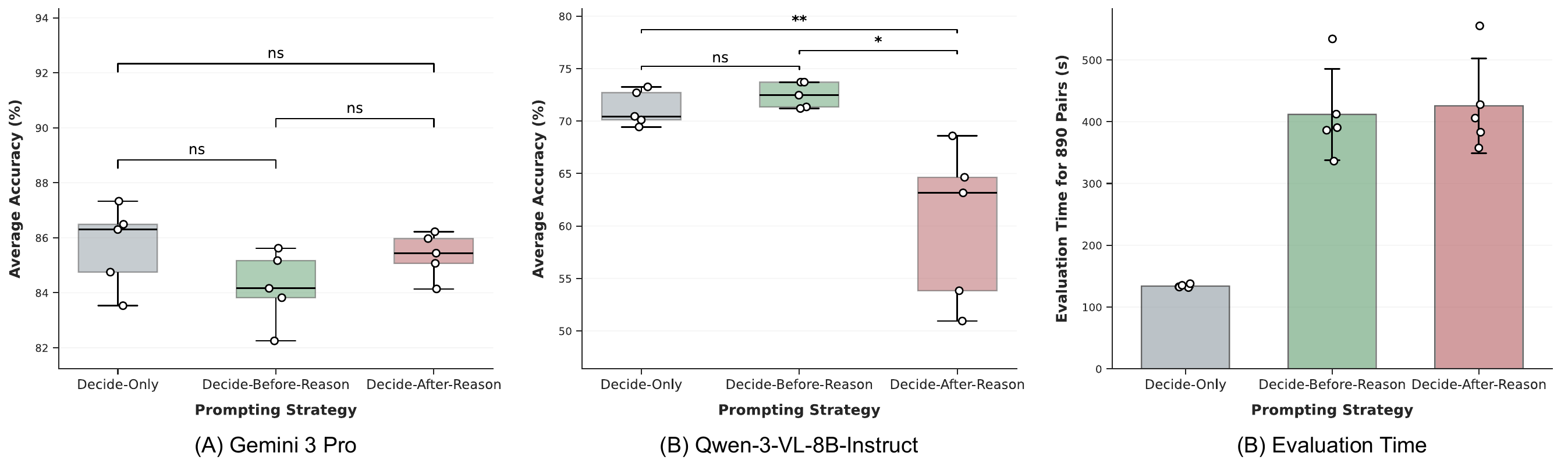}
    \caption{\textbf{Comparison results of three different prompting strategies} for visual consistency evaluation on EditReward-Bench~\citep{luo2025editscore}. (a) and (b) show the results of Gemini 3 Pro and Qwen3-VL-8B-Instruct, respectively. Pairwise statistical significance is evaluated using the two–sided Mann–Whitney U test ($**p < 0.01$, $*p < 0.05$, ns: $p \geq 0.05$). (c) reports the average evaluation time over 890 testing pairs measured with Qwen3-VL-8B-Instruct.}
    \label{fig:prompt_strategy}
\end{figure}

\subsection{Pairwise Evaluation Prompts}\label{subsec:prompts}
This subsection introduces the pairwise evaluation prompts utilized by GEditBench v2 to rigorously evaluate the editing models from the Instruction Following, Visual Quality, and Visual Consistency dimensions.
In general, the evaluation prompts can be categorized into three distinct strategies based on their output format: (1) \textbf{Decide-Only} template forces the model to output the winning candidate immediately without any explanation;
(2) \textbf{Decide-After-Reason} template~\citep{luo2025editscore} requires the model to generate a detailed textual analysis before declaring the final winner;
and (3) \textbf{Decide-Before-Reason} template~\citep{ye2025unicedit} instructs the model to state the winner upfront and subsequently append a textual justification.
For accurate evaluation, we conducted the following comparative experiments on the visual consistency sub-set of EditReward-Bench~\citep{luo2025editscore}.

\noindent\textbf{Which Output Format Performs Better?}
We initially drafted a basic prompt for each of these three prompting strategies.
Then, we leverage Gemini~\citep{team2023gemini}, GPT~\citep{gpt4o20250325}, DeepSeek~\citep{guo2025deepseek}, and Doubao~\citep{seed1d8} to refine these prompts to obtain the other four unique prompt variations for each strategy.
Fig.~\ref{fig:prompt_strategy}~(a)~and~(b) show our experimental results across Gemini 3 Pro~\citep{team2023gemini} and Qwen3-VL-8B-Instruct~\citep{qwen3_vl}.
These results demonstrate marginal accuracy deviations across all three paradigms for the closed-source Gemini 3 Pro.
We attribute this robust performance to its superior multi-modal understanding and long context window.
Conversely, forcing the Qwen3-VL-8B-Instruct model to generate a rationale first triggers a significant performance degradation.
We attribute this failure to hallucinations in vision-language models, which arise because generating lengthy prior text dilutes core visual attention and corrupts the final structural judgment~\citep{jiang2025devils}.
Since the strategies ``Decide-Only'' and ``Decide-Before-Reason'' exhibit nearly identical accuracy for Qwen3-VL-8B-Instruct, we further record their evaluation time for a total of 890 testing pairs (see Fig.~\ref{fig:prompt_strategy}~(c)).
The result shows that the ``Decide-Only'' strategy substantially reduces the average evaluation time compared to the ``Decide-Before-Reason'' approach (133.93s vs 411.84s).

Therefore, we adopt the efficient ``Decide-Only'' prompting strategy for all three dimensions evaluation, and the prompts are detailed in Fig.~\ref{fig:if_prompt}, Fig.~\ref{fig:vq_prompt}, and Fig.~\ref{fig:vc_prompt}, respectively.

\section{Pairwise Visual Consistency Judge}\label{sec:pvc_judge}
\subsection{\texttt{(task, Pipeline, Region-Specific Metrics)} Mapping}\label{subsec:pipeline}
\begin{table}[t]
  \caption{\textbf{\texttt{(Sub-Task, Pipeline, Region-Specific Metrics)} mapping for preference pair automated annotation.} Primary metrics are marked in \textbf{bold}.}
  \label{tab:pipeline_map}
  \centering
  \resizebox{\linewidth}{!}{
  \begin{tabular}{lcccc}
    \toprule
    \multirow{2}{*}{\textbf{Sub-Task}} & \multirow{2}{*}{\textbf{Pipeline Type}} & \multirow{2}{*}{\textbf{Grounding Image}} & \multicolumn{2}{c}{\textbf{Region-Specific Metrics}} \\
    & & & Edit Region ($\Omega_{edit}$) & Non-edit Region ($\Omega_{non}$) \\
    \midrule
    subject addition & object-centric & output & $-$ & SSIM, \textbf{LPIPS}, EMD \\
    subject removal & object-centric & input & $-$ & SSIM, \textbf{LPIPS}, EMD \\
    subject replace & object-centric & input,output & $-$ & SSIM, \textbf{LPIPS}, EMD \\
    size adjustment & object-centric & input,output & \makecell{SAM-based CLIP \texttt{[CLS]} similarity, \\\textbf{SAM-based DINO \texttt{[CLS]} similarity}} & \textbf{LPIPS}, EMD \\
    color alteration & object-centric & input,output & \makecell{\textbf{L-channel SSIM}, \\DINO structure similarity} & \textbf{LPIPS}, EMD \\
    material modification & object-centric & input,output & \makecell{\textbf{depth SSIM}, \\DINO structure similarity} & \textbf{LPIPS}, EMD \\
    portrait beautification & human-centric & input,output & \makecell{\textbf{Face ID}, Hair Appearance, \\Body Appearance} & \makecell{\textbf{LPIPS}, \\BG Face ID similarity} \\
    motion change & human-centric & input,output & \makecell{\textbf{Face ID}, Hair Appearance, \\Body Appearance} & \makecell{\textbf{LPIPS}, \\BG Face ID similarity} \\
    text editing & object-centric & input,output & $-$ & SSIM, \textbf{LPIPS}, EMD \\
    character reference & human-centric & input & \textbf{max match Face ID} & $-$ \\
    object reference & object-centric & input,output & \makecell{SAM-based CLIP \texttt{[CLS]} similarity, \\\textbf{SAM-based DINO \texttt{[CLS]} similarity}} & $-$ \\
  \bottomrule
  \end{tabular}
  }
\end{table}
We develop object- and human-centric automated pipelines to efficiently construct high-quality preference pairs.
Our method dynamically partitions both input and output images into edit and non-edit regions, and applies dedicated metrics to each region to assess overall visual consistency. Table~\ref{tab:pipeline_map}~provides a comprehensive mapping between tasks, pipeline types, grounding image, and the corresponding regional metrics.
Beyond SSIM~\citep{wang2004image} and LPIPS~\citep{zhang2018unreasonable}, we further incorporate the following metrics:
\begin{itemize}
    \item \textbf{EMD}: We first extract patch-level embeddings from the non-edit regions using the CLIP vision encoder~\citep{radford2021clip}, and then compute the Earth Mover’s Distance~\citep{rubner1998metric} between the input and output feature distributions to quantify semantic preservation.

    \item \textbf{SAM-based CLIP \texttt{[CLS]} similarity}:
    We first isolate the target foreground object using the Segment Anything Model to eliminate background interference.
    The global \texttt{[CLS]} token is then extracted from the CLIP vision encoder to compute a category-level cosine similarity between input and output, which helps mitigate consistency artifacts caused by viewpoint changes or size variations.

    \item \textbf{SAM-based DINO \texttt{[CLS]} similarity}: Similar to SAM-based CLIP \texttt{[CLS]} similarity, except that the global \texttt{[CLS]} token is extracted from the DINOv3~\citep{simeoni2025dinov3} to compute the cosine similarity.

    \item \textbf{L-channel SSIM}: Designed for the delicate color alteration task.
    To prevent legitimate color edits from artificially lowering the consistency score, we decouple structural integrity from chromatic variations.
    Specifically, we extract only the L (Lightness) channel from the edit region and compute the SSIM on this illumination map, enabling an unbiased evaluation of structural consistency.

    \item \textbf{DINO Structure Similarity}: We extract patch embeddings from the edit region using DINOv3, and construct spatial self-similarity matrices for both the input and output images.
    The $L_2$ distance between these matrices is then computed to measure structural consistency, while remaining robust to surface-level texture variations.

    \item \textbf{depth SSIM}: Since material changes often alter surface illumination and distort the lightness channel, we instead focus on geometric consistency.
    We use the Depth Anything V2 model~\citep{yang2024depth} to extract depth maps from both the input and output images, and compute the SSIM score between them within the edit region. This geometry-based metric evaluates shape preservation independently of material appearance.

    \item \textbf{Face ID}: We use the ArcFace~\citep{deng2019arcface} network to extract identity embeddings from the localized target face in both the input and output images, and compute their cosine similarity to verify that the generated result preserves the subject's facial identity.

    \item \textbf{Hair Appearance}: A dedicated hair segmenter is first applied within the edit region to obtain binary masks for both the input and output images.
    We then extract high-frequency texture maps by subtracting a Gaussian-blurred version from the masked crop:$H {=} I_{hair} {-} (G_{\sigma} {*} I_{hair}),$ where $I_{hair}$ denotes the isolated hair region and $G_{\sigma}$ is a Gaussian kernel with standard deviation $\sigma$.
    The absolute difference between the paired high-frequency maps is finally computed to measure the microscopic structural consistency of hair strands.

    \item \textbf{Body Appearance}: Specifically evaluates clothing and pose consistency by isolating the human torso and limbs.
    A selfie segmenter is first applied to obtain the full human silhouette and remove background interference.
    The previously detected hair mask and facial bounding box are then subtracted to produce a mask containing only the body region.
    Dense patch embeddings are extracted within this mask using the DINOv3 and spatially averaged to form a unified semantic representation.
    The cosine similarity between the averaged embeddings of the input and output images is finally computed to measure body fidelity.

    \item \textbf{BG Face ID similarity}: Specifically targets attribute leakage during human-centric edits, where instructions may spill over to non-target background individuals and degrade global consistency.
    Face detectors are first applied within the non-edit region to extract the background faces' bounding boxes in both the input and output images.
    Intersection over Union (IoU) is used to establish correspondences between these paired faces.
    For each matched pair, identity embeddings are extracted using the ArcFace network, and cosine similarity is computed.
    The final score is obtained by averaging the similarities across all background faces to measure identity preservation for non-target individuals.

    \item \textbf{max match Face ID}: Specifically targets the unconstrained character reference task, where large spatial transformations between the input reference and the synthesized output make positional tracking unreliable.
    We first extract the target subject's identity embedding by the ArcFace network from the edit region of the input image.
    Face detectors are then applied to the output image to obtain identity embeddings for all generated faces.
    Cosine similarity is computed between the reference embedding and each candidate, and the maximum score is returned.
    This global matching strategy enables reliable identity verification under large positional or scale variations.
\end{itemize}

\begin{figure}[!t]
    \centering
    \includegraphics[width=.8\linewidth]{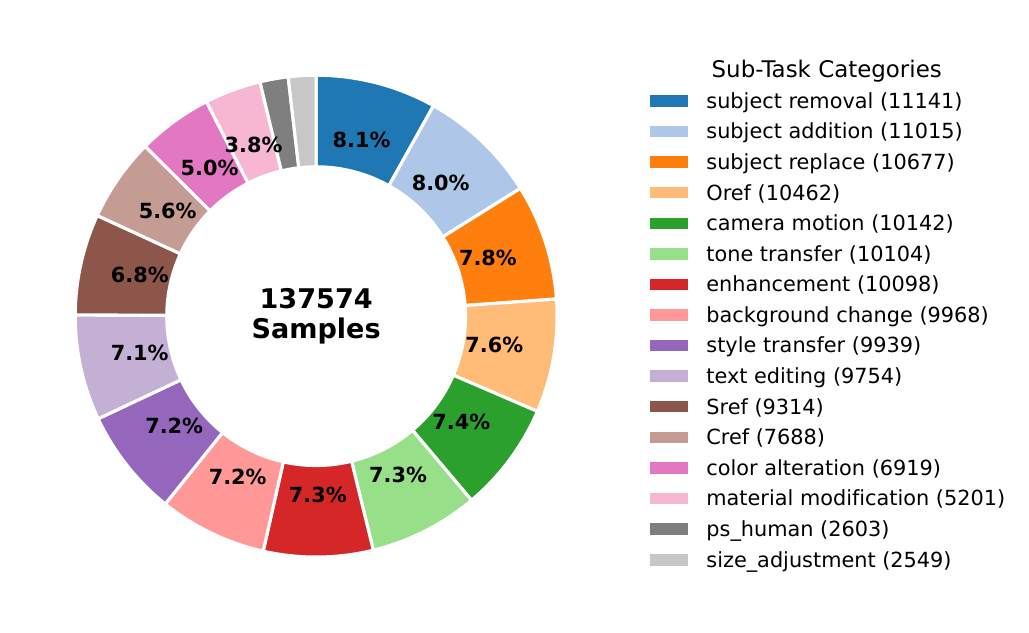}
    \caption{\textbf{Distribution of the preference dataset.}
    The donut chart shows the detailed allocation across 16 editing tasks, providing reliable signals for training the PVC-Judge model.
    }
    \label{fig:training_data_distri}
\end{figure}

\subsection{Preference Data Distribution}\label{subsec:data_distri}
The statistical distribution of the constructed preference dataset is shown in Fig.~\ref{fig:training_data_distri}. This balanced distribution provides robust and high-quality preference signals for training our PVC-Judge model.

\subsection{Training Hyper-parameters}\label{subsec:train_hyper}
Table~\ref{tab:train_param} summarizes the hyper-parameters for the LoRA model trained during our experiments.

\begin{table}[bh]
  \caption{\textbf{LoRA Training Parameters.}}
  \label{tab:train_param}
  \centering
  \begin{tabular}{lc}
    \toprule
    \textbf{Hyper-parameter} & \textbf{Value} \\
    \midrule
    optimizer & AdamW \\
    num\_train\_epochs & 3 \\
    per\_device\_train\_batch\_size & 2 \\
    gradient\_accumulation\_steps & 1 \\
    learning\_rate & 2.0e-6 \\
    warmup\_ratio & 0.05 \\
    lr\_scheduler\_type & cosine \\
    weight\_decay & 0.1 \\
    lora\_rank & 64 \\
    lora\_alpha & 128 \\
    lora\_dropout & 0.05 \\
    lora\_namespan\_exclude & [`lm\_head', `embed\_tokens'] \\
    image\_min\_pixels & $256*32*32$ \\
    image\_max\_pixels & $1280*32*32$ \\
  \bottomrule
  \end{tabular}
\end{table}

\section{Annotation Protocol for VCReward-Bench}\label{sec:vcReward_bench}

\begin{figure}[tb]
    \centering
    \includegraphics[width=1\textwidth]{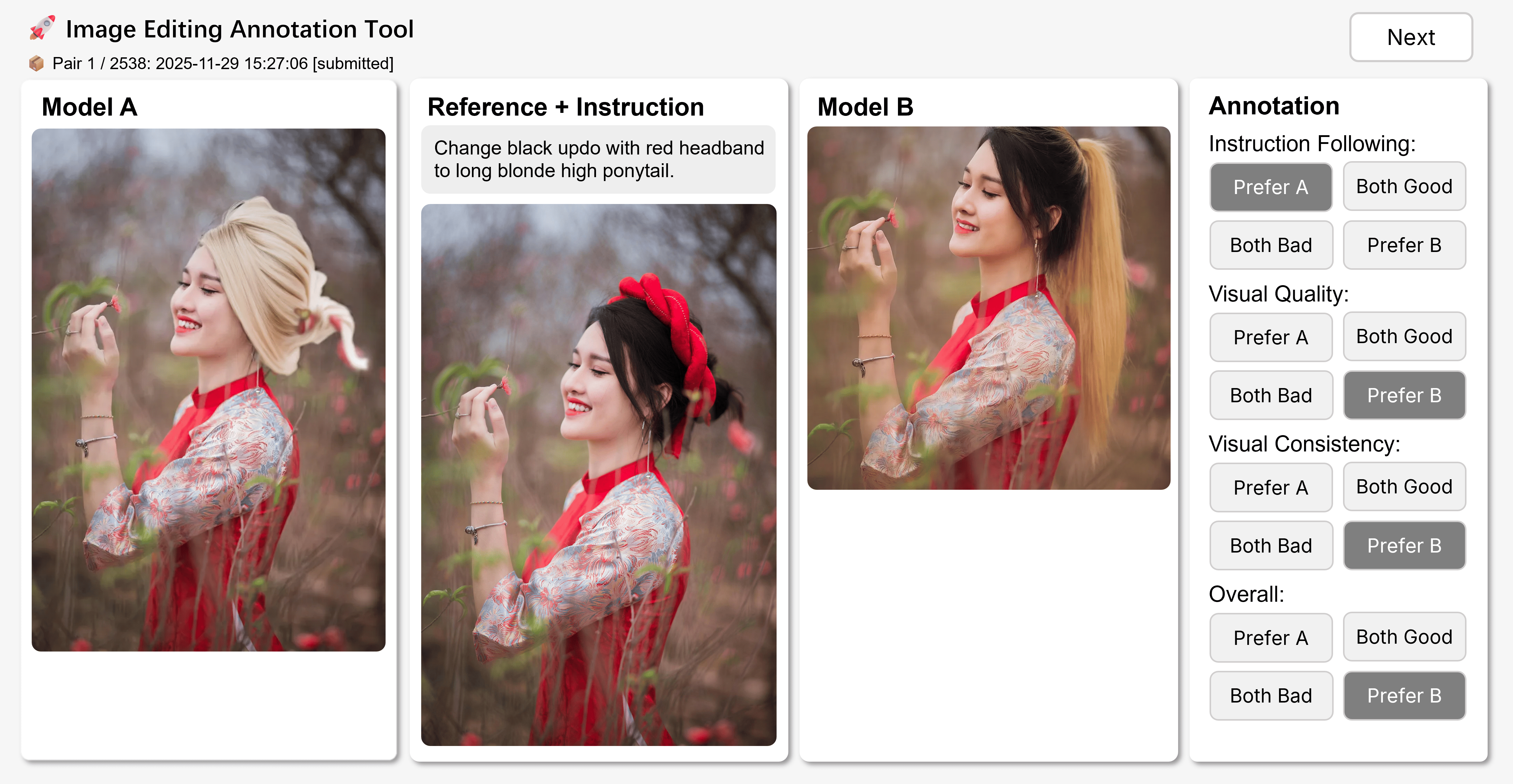}
    \caption{\textbf{Screenshots of our custom-built annotation interface for VCReward-Bench.}}
    \label{fig:reward_anno}
\end{figure}

\begin{figure}[tb]
    \centering
    \includegraphics[width=1\textwidth]{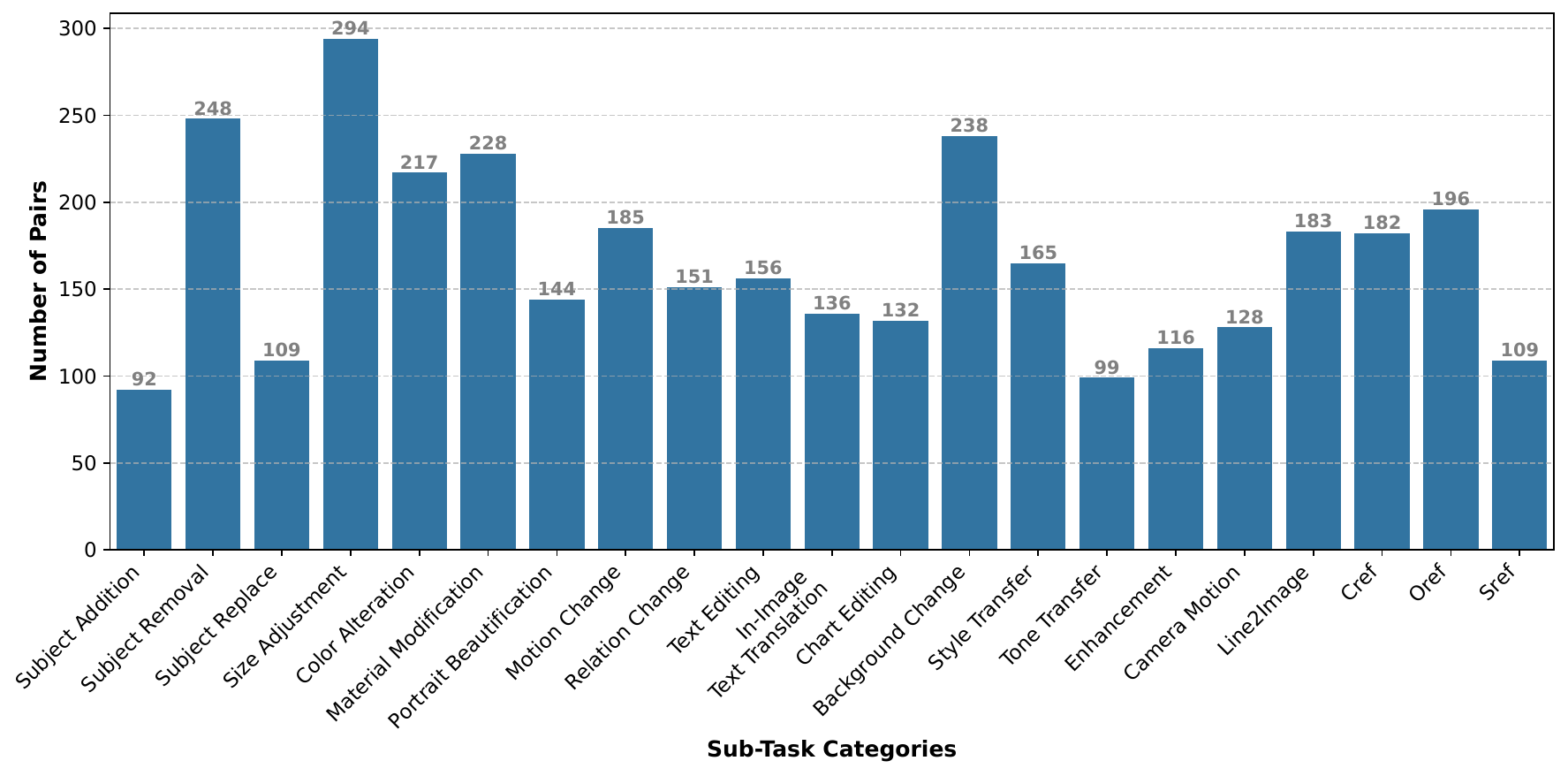}
    \caption{\textbf{Distribution of VCReward-Bench instances by tasks.}}
    \label{fig:VCReward_distri}
\end{figure}

To guarantee the reliability, diversity, and challenge level of VCReward-Bench, we designed a strict multi-model generation and filtering pipeline.
Initially, to capture a comprehensive distribution of editing behaviors, artifacts, and failure modes, we utilized a diverse ensemble of both leading open-source~\citep{yin2025reasonedit,wu2025qwenimagetechnicalreport,labs2025kontext} and proprietary models~\citep{gpt4o20250325,team2023gemini} to generate a vast pool of candidate images for each editing prompt.

These generated candidates were then formulated into pairwise comparisons.
Expert annotators evaluated each pair across three orthogonal dimensions and one global preference: \textit{Instruction Following} (IF), \textit{Visual Quality} (VQ), \textit{Visual Consistency} (VC), and \textit{Overall}.
The main interface is presented in Fig.~\ref{fig:reward_anno}.
This interactive interface independently isolates the decision process for every single evaluation dimension, ensuring a rigorous multi-dimensional assessment paradigm.
Furthermore, the system restricts the selection space to four options: ``Prefer A'', ``Both Good'', ``Both Bad'', and ``Prefer B''. The inclusion of explicit tie options helps absorb inevitable subjective variation among annotators, improving the statistical robustness of the evaluation.

For examining a model's ability to assess visual consistency rather than detect trivial prompt mismatches, we adopt a Pareto-style filtering strategy over multiple evaluation dimensions.
Let $\succ_d$ denote the preference relation under evaluation dimension $d$. For two edited images $I_A$ and $I_B$, the relation $I_A \succ_d I_B$ indicates that $I_A$ is preferred to $I_B$ with respect to dimension $d$.
A candidate pair $(I_A, I_B)$ is included in VCReward-Bench only if the following condition holds:
\[
I_A \succ_{\mathrm{VC}} I_B, \quad
I_A \succeq_d I_B, \; \forall d \in \{\mathrm{IF, VQ, Overall}\}.
\]
That is, $I_A$ must be strictly preferred to $I_B$ on the visual consistency dimension, while being non-inferior across all remaining evaluation dimensions. This filtering procedure removes trivial pairs and ensures that retained pairs primarily differ in visual consistency, providing more informative signals for meta-evaluation.

Finally, this process resulted in a large-scale, high-quality dataset of 3,508 testing preference pairs spanning 21 tasks, with the detailed distribution shown in Fig.~\ref{fig:VCReward_distri}.

\section{Additional Results}\label{sec:add_results}
\subsection{Full Numerical Results of Meta-Evaluation}\label{subsec:add_numerial}
Tables~\ref{tab:editscore_bench} and~\ref{tab:vcreward_bench} report the numerical meta-evaluation performance of various assessment models on the visual consistency dimension for EditReward-Bench~\citep{luo2025editscore} and VCReward-Bench.
The best-performing open-source model is highlighted in bold for easy identification.
Our proposed PVC-Judge consistently leads the open-source models, achieving state-of-the-art results across most complex tasks.
Moreover, it demonstrates assessment capabilities comparable to large closed-source proprietary models.

\begin{table}[tb]
  \caption{\textbf{Numerical results of assessment models for visual consistency on EditReward-Bench}~\citep{luo2025editscore}. \textbf{Bold} marks the best-performing open-source model.}
  \label{tab:editscore_bench}
  \centering
  \resizebox{\linewidth}{!}{
  \begin{tabular}{lccccc>{\columncolor{gray!20}}c}
    \toprule
    \multirow{2}{*}{\textbf{Sub-Tasks}} & \multirow{2}{*}{\textbf{Gemini 3 Pro}} & \multirow{2}{*}{\textbf{GPT 5.1}} & \textbf{EditReward} & \textbf{EditScore Avg@4} & \multirow{2}{*}{\textbf{Qwen3-VL-8B-Instruct}} & \multirow{2}{*}{\textbf{PVC-Judge}} \\
    & & & \citep{wu2025editreward} & \citep{luo2025editscore} & & \\
    \midrule
    Subject Addition & 80.88 & 80.88 & 74.38 & 57.65 & 72.06 & \textbf{77.94} \\
    Subject Removal & 83.56 & 63.01 & \textbf{89.90} & 51.91 & 61.64 & 67.12 \\
    Subject Replace & 90.57 & 77.36 & 81.55 & 54.85 & 73.58 & \textbf{84.91} \\
    Color Alteration & 91.30 & 86.96 & 83.49 & 77.83 & 72.46 & \textbf{94.20} \\
    Material Modification & 90.16 & 88.52 & 67.00 & 51.23 & 75.41 & \textbf{83.61} \\
    Style Change & 88.89 & 87.50 & 74.39 & 61.38 & 76.39 & \textbf{86.11} \\
    Tone Transfer & 86.67 & 83.33 & 72.61 & 55.65 & 78.33 & \textbf{80.00} \\
    Background Change & 92.00 & 83.00 & 77.15 & 63.25 & 74.00 & \textbf{87.00} \\
    Extract & 78.95 & 50.88 & \textbf{93.66} & 76.06 & 56.14 & 71.93 \\
    Portrait Beautification & 94.12 & 89.71 & 69.02 & 29.41 & 85.29 & \textbf{89.71} \\
    Text Modification & 84.93 & 75.34 & 84.38 & 51.56 & 65.75 & \textbf{87.67} \\
    Motion Change & 87.84 & 75.68 & 75.83 & 58.75 & 74.32 & \textbf{82.43} \\
    Hybrid & 85.48 & 80.65 & 75.00 & 62.26 & 74.19 & \textbf{79.03} \\
    \midrule
    Avg. Accuracy & 87.33 & 78.68 & 78.34 & 57.83 & 72.27 & \textbf{82.44} \\
  \bottomrule
  \end{tabular}
  }
\end{table}
\begin{table}[tb]
  \caption{\textbf{Numerical results of assessment models for visual consistency on our VCReward-Bench.} \textbf{Bold} marks the best-performing open-source model.}
  \label{tab:vcreward_bench}
  \centering
  \resizebox{\linewidth}{!}{
  \begin{tabular}{lccccc>{\columncolor{gray!20}}c}
    \toprule
    \textbf{Sub-Tasks} & \textbf{Gemini 3 Pro} & \textbf{GPT 5.1} & \textbf{EditReward}~\cite{wu2025editreward} & \textbf{EditScore Avg@4}~\cite{luo2025editscore} & \textbf{Qwen3-VL-8B-Instruct} & \textbf{PVC-Judge} \\
    \midrule
    Subject Addition & 93.59 & 80.68 & 67.39 & 41.30 & 72.53 & \textbf{86.96} \\
    Subject Removal & 86.32 & 68.49 & 70.97 & 35.48 & 69.01 & \textbf{74.19} \\
    Subject Replace & 85.88 & 67.96 & 65.14 & 36.70 & 74.00 & \textbf{82.57} \\
    Size Adjustment & 83.52 & 69.45 & 57.34 & 34.81 & 68.22 & \textbf{72.35} \\
    Color Alteration & 92.90 & 75.96 & 75.46 & 53.24 & 67.01 & \textbf{87.04} \\
    Material Modification & 86.91 & 80.09 & 57.02 & 50.88 & 72.89 & \textbf{86.84} \\
    Portrait Beautification & 81.42 & 70.99 & 59.03 & 34.03 & \textbf{76.23} & 75.69 \\
    Motion Change & 85.06 & 79.89 & 57.84 & 37.84 & 72.93 & \textbf{84.32} \\
    Relation Change & 85.94 & 72.97 & 81.46 & 50.99 & 71.33 & \textbf{82.78} \\
    Text Editing & 87.83 & 79.33 & 76.92 & 50.00 & 77.87 & \textbf{78.21} \\
    In-Image Text Translation & 96.72 & 83.97 & \textbf{90.44} & 58.82 & 83.09 & 85.29 \\
    Chart Editing & 94.66 & 92.25 & \textbf{92.42} & 82.58 & 87.60 & 89.39 \\
    Background Change & 87.07 & 80.08 & 53.36 & 50.84 & 72.69 & \textbf{86.13} \\
    Style Transfer & 84.08 & 82.91 & 56.36 & 41.21 & 73.62 & \textbf{84.24} \\
    Tone Transfer & 86.60 & 85.57 & 49.49 & 65.66 & 78.79 & \textbf{86.87} \\
    Enhancement & 85.05 & 72.41 & 68.97 & 57.76 & 69.83 & \textbf{80.17} \\
    Camera Motion & 88.28 & 82.54 & 60.94 & 57.03 & 76.56 & \textbf{79.69} \\
    Line2Image & 76.27 & 68.85 & 56.83 & 31.69 & 51.37 & \textbf{66.12} \\
    Cref & 81.01 & 65.68 & 63.19 & 48.35 & 68.68 & \textbf{80.22} \\
    Oref & 89.42 & 77.86 & 67.86 & 56.12 & 68.97 & \textbf{81.12} \\
    Sref & 91.18 & 76.74 & 87.16 & 57.80 & 81.31 & \textbf{88.07} \\
    \midrule
    Avg. Accuracy & 87.13 & 76.89 & 67.41 & 49.20 & 73.07 & \textbf{81.82} \\
  \bottomrule
  \end{tabular}
  }
\end{table}

\subsection{Qualitative Analysis}\label{subsec:add_qualitative}
\begin{figure}[thb]
    \centering
    \includegraphics[width=1\linewidth]{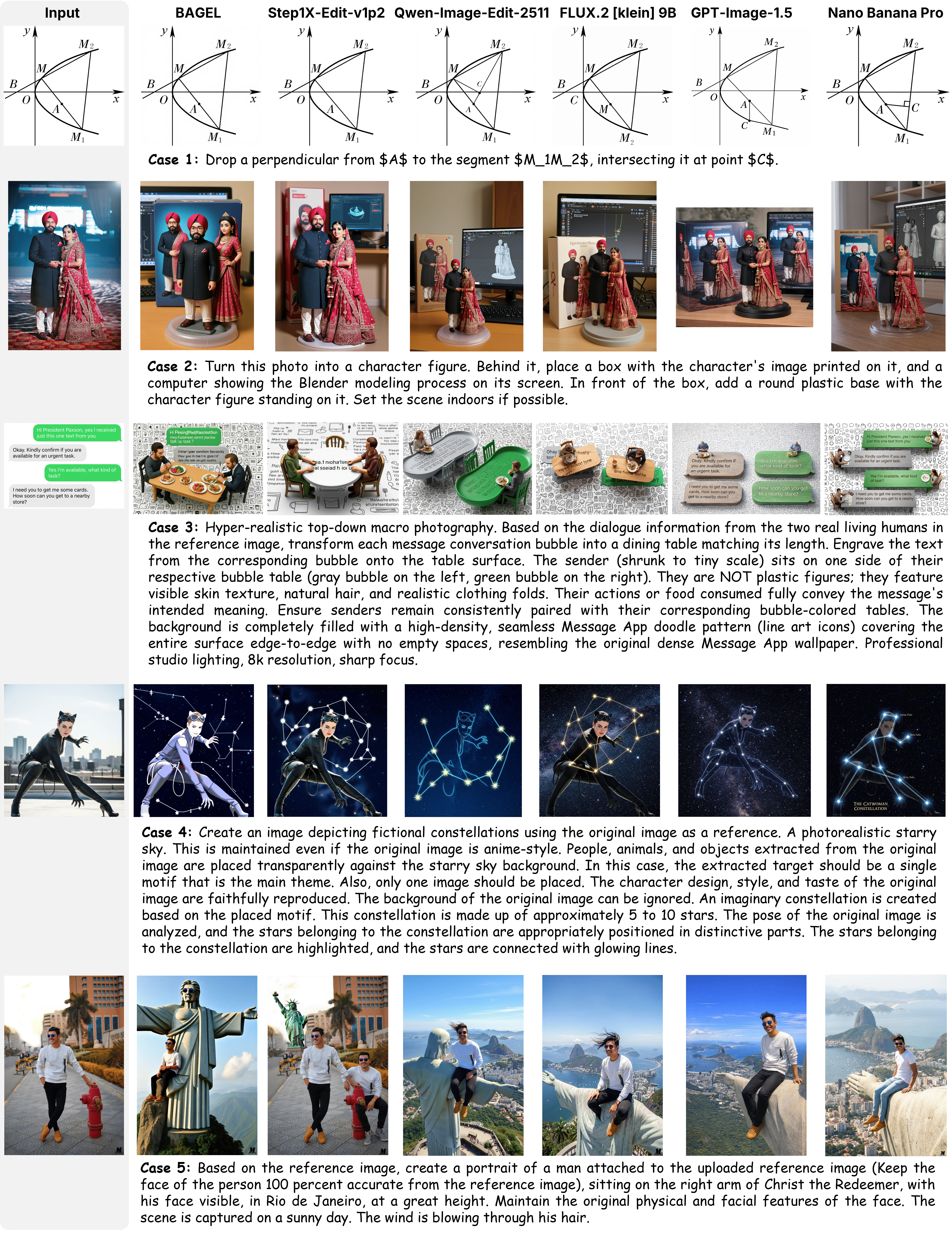}
    \caption{\textbf{Open-set editing examples across six representative models.} Many open-source models fail to fully capture implicit user instructions, leading to partial or inconsistent edits. This highlights the need for open-set evaluation to reveal limitations of current generative editing models.}
    \label{fig:openset_ana1}
\end{figure}

\noindent\textbf{Open-Set Task.}
Fig.~\ref{fig:openset_ana1} comprehensively visualizes diverse cases across six representative editing models.
One could observe that current open-source models exhibit difficulty in interpreting implicit user intentions, often overlooking subtle textual constraints contained in unpredictable prompts.
These qualitative observations highlight a key limitation: models trained on predefined categories may not generalize reliably to open-set instructions.
Furthermore, errors frequently manifest as missing or misaligned edits, inconsistent object attributes, or partial adherence to the user prompt, suggesting that current models still struggle to robustly fuse complex textual cues with visual generation.
This analysis underscores the importance of evaluating visual editing performance beyond narrowly pre-defined tasks and demonstrates the utility of open-set benchmarks in revealing real-world limitations of generative editing models.

\begin{figure}[thb]
    \centering
    \includegraphics[width=1\linewidth]{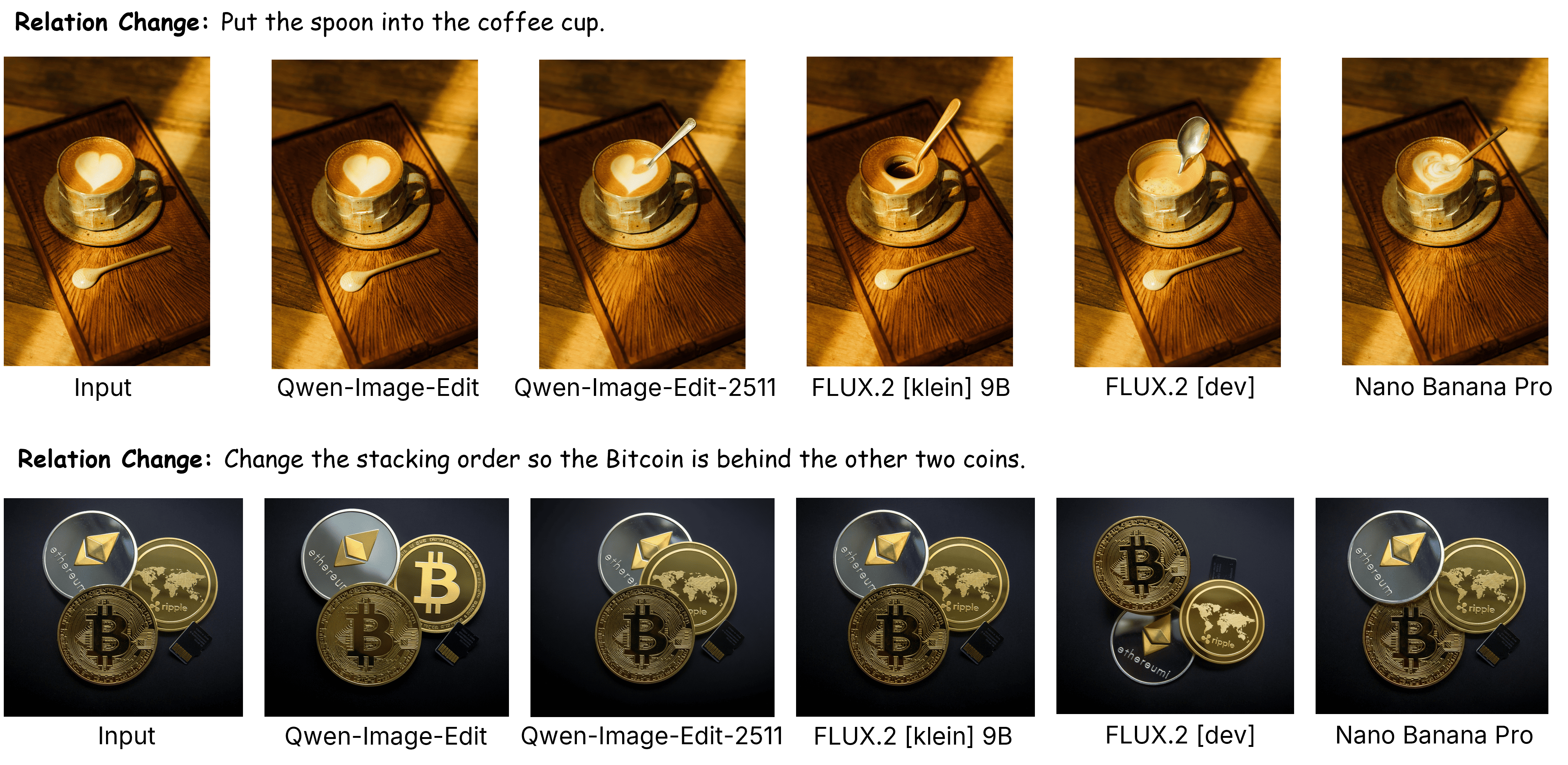}
    \caption{\textbf{Visualization results of relation change task.} Open-source models often struggle to capture complex spatial relationships, leading to partial or simplified relational edits, while closed-source models better preserve inter-object dependencies.}
    \label{fig:relation_ana}
\end{figure}

\noindent\textbf{Weak Perception of Inter-Object Relations.}
Fig.~\ref{fig:relation_ana} presents generative results of representative models on the relation change task.
Visual comparisons reveal clear gaps between current open-source models and stronger closed-source counterparts.
Open-source models often struggle to accurately capture spatial relationships between objects, reducing complex relational edits to simple operations such as disjointed object addition or removal.
In some cases, models partially or entirely ignore the provided spatial instructions, highlighting limitations in handling structured inter-object dependencies.

\begin{figure}[thb]
    \centering
    \includegraphics[width=1\linewidth]{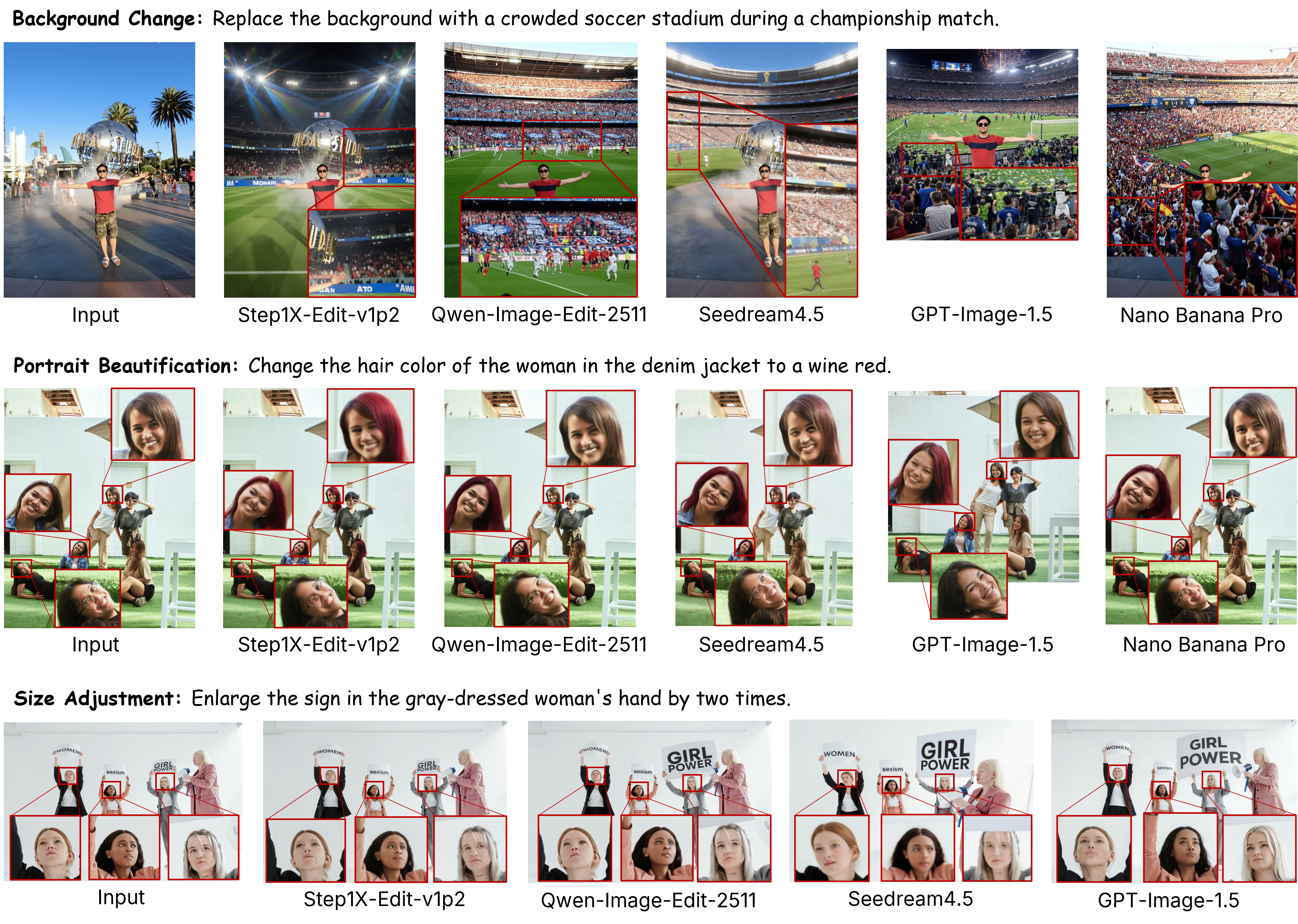}
    \caption{\textbf{Fine-grained detail generation across multiple tasks.} Open-source models frequently distort small entities and background subjects, whereas closed-source models better maintain structure, with some difficulty preserving identity on small faces.}
    \label{fig:small_face}
\end{figure}

\noindent\textbf{Struggle with Small Faces.}
Fig.~\ref{fig:small_face} shows the results of editing models in generating fine-grained details.
We observe that open-source models often produce distorted structures on small or background subjects, with facial regions being particularly challenging.
Closed-source models alleviate these structural issues but struggle to maintain identity consistency on small faces (e.g., GPT-Image-1.5).
These observations reveal a persistent gap in detail fidelity, which remains a key challenge for the practical deployment of generative editing models.

\begin{figure}[htbp]
    \centering
    \begin{JudgeBox}{Dimension I: Instruction Following}
    \textbf{\textcolor{blue!70!black}{[SYSTEM PROMPT]}} \\
    You are a meticulous instruction compliance analyst. Your sole purpose is to evaluate images based on how strictly they adhere to a given text instruction. You must be objective and focus exclusively on the fulfillment of the user's command. Output your answer in a structured JSON format.
    
    \tcbline
    
    \textbf{\textcolor{orange!80!black}{[USER PROMPT]}} \\
    **Task: Evaluate Instruction Adherence ONLY.**

    You must determine which of the two generated images, Image A or Image B, more accurately and completely follows the editing instruction, based on the original image.
    \\
    
    **Your Single Focus:**\\
    - Did the image edit the correct object or region?\\
    - Did the image apply the correct style, color, or content as requested?\\
    - Were all parts of the instruction addressed?\\

    **Your Thought Process:**\\
    1. Analyze Image A's adherence to the instruction.\\
    2. Analyze Image B's adherence to the instruction.\\
    3. Compare them and declare a winner based *only* on adherence.\\
    4. Provide a winner in the specified JSON format.\\

    **Required Output JSON Schema:**\\
    \textasciigrave\textasciigrave\textasciigrave json\\
    \{\\
        \hspace*{1em} ``winner'': `` Choose one from: `Image A', `Image B', or `Tie' ''\\
    \}\\
    \textasciigrave\textasciigrave\textasciigrave \\

    **Editing Instruction:** ``\textless instruction\textgreater''\\
    Original Image: \textless image\_1\textgreater \\

    -{}-{}- Generated Images to Compare -{}-{}-\\
    Image A: \textless image\_2\textgreater \\
    Image B: \textless image\_3\textgreater
\end{JudgeBox}
    \caption{The pairwise evaluation prompt for the \textbf{Instruction Following} dimension.}
    \label{fig:if_prompt}
\end{figure}
\begin{figure}[htbp]
    \centering
    \begin{LongJudgeBox}{Dimension II: Visual Quality}
    \textbf{\textcolor{blue!70!black}{[SYSTEM PROMPT]}} \\
    You are a strict and highly critical image quality comparator. Your sole responsibility is to determine which image has better intrinsic visual quality. You must evaluate only what is visually observable in the two images.\\
    
    Your judgment must be based exclusively on:\\
    1) Objective visual defects and logical plausibility.\\
    2) Visual aesthetics and artistic quality.\\

    You must prioritize the absence of defects and logical errors over artistic appeal.
    Be decisive. Be critical. Avoid being overly generous.\\

    Output only the required JSON. Do not output explanations.

    \tcbline
    
    \textbf{\textcolor{orange!80!black}{[USER PROMPT]}} \\
    Task: Decide which image has better overall visual quality.\\
    You are given two generated images:\\
    - Image A\\
    - Image B\\
    You must compare the two images purely based on their intrinsic quality.\\
    
    -{}-{}-{}-{}-{}-{}-{}-{}-{}-{}-{}-{}-{}-{}-{}-{}-{}-{}-{}-{}-{}-{}-{}-{}-{}-{}-{}-{}-{}-{}-{}-{}-{}-{}-{}-{}-{}-{}-{}-{}-{}-{}-{}-{}-{}-{}-{}-{}-{}-\\
    Evaluation Dimension I: Artifacts \& Logical Flaws (Higher Priority)\\
    -{}-{}-{}-{}-{}-{}-{}-{}-{}-{}-{}-{}-{}-{}-{}-{}-{}-{}-{}-{}-{}-{}-{}-{}-{}-{}-{}-{}-{}-{}-{}-{}-{}-{}-{}-{}-{}-{}-{}-{}-{}-{}-{}-{}-{}-{}-{}-{}-{}-\\
    Carefully inspect both images for objective defects:\\
    • Anatomical abnormalities (extra fingers, distorted limbs, fused body parts)\\
    • Physically impossible structures\\
    • Broken shadows or inconsistent lighting direction\\
    • Incorrect reflections in mirrors or water\\
    • Perspective inconsistencies\\
    • Floating or unsupported objects\\
    • Garbled or unreadable text\\
    • Background warping or structural distortion\\
    • Texture melting or unnatural blending\\
    • Over-smoothing (plastic-like skin) or unnatural high-frequency noise\\
    • Local blur patches or incoherent fine details\\
    
    If one image contains more severe or more frequent defects, the cleaner image should win.\\
    
    Major structural or logical flaws outweigh aesthetic advantages.\\
    
    -{}-{}-{}-{}-{}-{}-{}-{}-{}-{}-{}-{}-{}-{}-{}-{}-{}-{}-{}-{}-{}-{}-{}-{}-{}-{}-{}-{}-{}-{}-{}-{}-{}-{}-{}-{}-{}-{}-{}-{}-{}-{}-{}-{}-{}-{}-{}-{}-{}-\\
    Evaluation Dimension II: Visual Aesthetics \& Artistry\\
    -{}-{}-{}-{}-{}-{}-{}-{}-{}-{}-{}-{}-{}-{}-{}-{}-{}-{}-{}-{}-{}-{}-{}-{}-{}-{}-{}-{}-{}-{}-{}-{}-{}-{}-{}-{}-{}-{}-{}-{}-{}-{}-{}-{}-{}-{}-{}-{}-{}-\\
    If both images are similarly free of major defects, compare their artistic and perceptual quality:\\
    • Composition balance and clarity of subject\\
    • Visual hierarchy and focal emphasis\\
    • Color harmony and tonal coherence\\
    • Lighting depth and dimensionality\\
    • Realism of materials and textures\\
    • Sharpness and clarity of details\\
    
    -{}-{}-{}-{}-{}-{}-{}-{}-{}-{}-{}-{}-{}-{}-{}-{}-{}-{}-{}-{}-{}-{}-{}-{}-{}-{}-{}-{}-{}-{}-{}-{}-{}-{}-{}-{}-{}-{}-{}-{}-{}-{}-{}-{}-{}-{}-{}-{}-{}-\\
    Decision Rule\\
    -{}-{}-{}-{}-{}-{}-{}-{}-{}-{}-{}-{}-{}-{}-{}-{}-{}-{}-{}-{}-{}-{}-{}-{}-{}-{}-{}-{}-{}-{}-{}-{}-{}-{}-{}-{}-{}-{}-{}-{}-{}-{}-{}-{}-{}-{}-{}-{}-{}-\\
    1. Prefer the image with fewer and less severe defects.\\
    2. If both are similarly clean, prefer the image with superior aesthetics.\\
    3. If both images are extremely similar in overall quality, return "Tie".\\
    
    -{}-{}-{}-{}-{}-{}-{}-{}-{}-{}-{}-{}-{}-{}-{}-{}-{}-{}-{}-{}-{}-{}-{}-{}-{}-{}-{}-{}-{}-{}-{}-{}-{}-{}-{}-{}-{}-{}-{}-{}-{}-{}-{}-{}-{}-{}-{}-{}-{}-\\
    Required Output JSON\\
    -{}-{}-{}-{}-{}-{}-{}-{}-{}-{}-{}-{}-{}-{}-{}-{}-{}-{}-{}-{}-{}-{}-{}-{}-{}-{}-{}-{}-{}-{}-{}-{}-{}-{}-{}-{}-{}-{}-{}-{}-{}-{}-{}-{}-{}-{}-{}-{}-{}-\\
    \textasciigrave\textasciigrave\textasciigrave json\\
    \{\\
        \hspace*{1em} ``winner'': `` Choose one from: `Image A', `Image B', or `Tie' ''\\
    \}\\
    \textasciigrave\textasciigrave\textasciigrave \\

    -{}-{}- Generated Images to Compare -{}-{}-\\
    Image A: \textless image\_1\textgreater\\
    Image B: \textless image\_2\textgreater
\end{LongJudgeBox}
    \caption{The pairwise evaluation prompt for the \textbf{Visual Quality} dimension.}
    \label{fig:vq_prompt}
\end{figure}
\begin{figure}[htbp]
    \centering
    \begin{JudgeBox}{Dimension III: Visual Consistency}
    \textbf{\textcolor{blue!70!black}{[SYSTEM PROMPT]}} \\
    You are a meticulous Image Integrity Analyst. Your function is to detect unintended modifications by comparing a generated image to its original source. Your sole purpose is to assess whether regions *outside the explicit scope of the editing instruction* have been unintentionally altered. You must first deduce the intended target of the edit, then ignore that target area and focus exclusively on the preservation of all other regions. Output your answer in a structured JSON format.
    
    \tcbline
    
    \textbf{\textcolor{orange!80!black}{[USER PROMPT]}} \\
    **Task: Evaluate Preservation of Non-Target Regions ONLY.**\\

    You will be given an **Original Image**, an **Editing Instruction**, and two generated images: **Image A** and **Image B**. Your job is to determine which generated image did a better job of keeping the areas *not targeted by the instruction* identical to the **Original Image**.\\
    
    **Your Single Focus:**\\
    - **Scope Analysis:** First, what is the specific object, area, or concept the instruction intended to change?\\
    - **Collateral Impact on Style:** In areas outside the instruction's direct scope, did the overall artistic style, mood, or filter change compared to the original?\\
    - **Color \& Lighting Spillover:** Did the color grading, saturation, or lighting shift in regions that should not have been affected by the edit? Compare them strictly against the Original Image.\\
    - **Integrity of Untouched Elements:** Were any objects, patterns, or structures that were *not supposed to be edited* distorted, added, or removed?\\
    - **Texture \& Detail Fidelity:** In untouched parts of the image, was the original texture, sharpness, and fine detail preserved, or did it become blurry, artificially smooth, or noisy?\\
    
    **Your Thought Process:**\\
    1.  **Deconstruct the Instruction:** First, analyze the editing instruction to clearly define the intended target area(s) of the modification.\\
    2.  **Analyze Image A:** Compare the regions *outside* of this target area in Image A against the Original Image, looking for any unintended changes.\\
    3.  **Analyze Image B:** Compare the regions *outside* of this target area in Image B against the Original Image, looking for any unintended changes.\\
    4.  **Declare a Winner:** Decide which image better preserved the non-target regions with higher fidelity to the Original Image. The image with fewer or less severe unintended changes wins. Provide your winner in the specified JSON format.\\

    **Required Output JSON Schema:**\\
    \textasciigrave\textasciigrave\textasciigrave json\\
    \{\\
        \hspace*{1em} ``winner'': `` Choose one from: `Image A', `Image B', or `Tie' ''\\
    \}\\
    \textasciigrave\textasciigrave\textasciigrave \\

    **Editing Instruction:** ``\textless instruction\textgreater''\\
    
    Original Image: \textless image\_1\textgreater \\

    -{}-{}- Generated Images to Compare -{}-{}-\\
    Image A: \textless image\_2\textgreater \\
    Image B: \textless image\_3\textgreater
\end{JudgeBox}
    \caption{The pairwise evaluation prompt for the \textbf{Visual Consistency} dimension.}
    \label{fig:vc_prompt}
\end{figure}

\end{document}